\documentclass[journal,compsoc]{IEEEtran}
\usepackage{epsfig}
\usepackage{graphicx}
\usepackage{amsmath}
\usepackage{amssymb}
\usepackage{algorithm, algorithmic}

\usepackage{diagbox}
\usepackage{float}
\usepackage{afterpage}
\usepackage{bm}
\usepackage{subfig}

\usepackage{multirow}
\usepackage{color}
\usepackage{tablefootnote}
\usepackage{adjustbox}
\usepackage{wrapfig}

\usepackage{hyperref}       % hyperlinks
\usepackage{url}            % simple URL typesetting
\usepackage{booktabs}       % professional-quality tables
\usepackage{amsfonts}       % blackboard math symbols
\usepackage{nicefrac}       % compact symbols for 1/2, etc.
\usepackage{microtype}      % microtypography
\usepackage{times}
\usepackage{epsfig}

\usepackage{bbding}
\usepackage{etoolbox}
\usepackage{paralist}
\usepackage{ulem}
\usepackage{tikz}
\usepackage{color}

\usepackage{makecell}

\usepackage{xcolor,colortbl}

\newcolumntype{Y}{p{0.5cm}<{\centering}}

\definecolor{Gray}{gray}{0.5}
\definecolor{LightCyan}{rgb}{0.88,1,1}

\newcolumntype{a}{>{\columncolor{Gray}}c}
\newcolumntype{b}{>{\columncolor{white}}c}

\def\etal{\textit{et al}.}
\def\ie{\textit{i.e.}}
\def\eg{\textit{e.g.}}
\def\etc{\textit{etc}}
\def\wrt{\textit{w.r.t. }}

\newcommand{\tb}[1]{\textbf{#1}}
\newcommand{\bc}[1]{\textcolor[RGB]{192,0,0}{\text{#1}}}
\newcommand{\rc}[1]{\textcolor{blue}{\text{#1}}}
\newcommand{\bb}[1]{\textcolor[RGB]{192,0,0}{\textbf{#1}}}
\newcommand{\rb}[1]{\textcolor{blue}{\textbf{#1}}}

\begin{document}

\title{Rotation-Adaptive Point Cloud Domain Generalization via Intricate Orientation Learning}

\author{{Bangzhen~Liu,~Chenxi~Zheng,~Xuemiao~Xu,~Cheng Xu,~Huaidong~Zhang, \\ and~Shengfeng~He,~\IEEEmembership{Senior Member,~IEEE}}

\thanks{This work is supported by the China National Key R\&D Program (No. 2023YFE0202700, 2024YFB4709200), the Key-Area Research and Development Program of Guangzhou City (No. 2023B01J0022), the Guangdong Provincial Natural Science Foundation for Outstanding Youth Team Project (No. 2024B1515040010), the NSFC Key Project (No. U23A20391), the Guangdong Natural Science Funds for Distinguished Young Scholars (No. 2023B1515020097), the AI Singapore Programme under the National Research Foundation Singapore (No. AISG3-GV-2023-011), and the Lee Kong Chian Fellowships. (Bangzhen~Liu and Chenxi~Zheng contributed equally to this work.) (Corresponding authors: Xuemiao~Xu; Cheng~Xu.)
}
\thanks{Bangzhen Liu,~Chenxi~Zheng, Xuemiao Xu, Cheng Xu, and Huaidong Zhang are with the South China University of Technology, Guangzhou, China.~E-mail: liubz.scut@gmail.com,~cszcx@mail.scut.edu.cn, xuemx@scut.edu.cn, cschengxu@gmail.com, and huaidongz@scut.edu.cn. Xuemiao Xu is also with the Guangdong Engineering Center for Large Model and GenAI Technology, the State Key Laboratory of Subtropical Building and Urban Science, the Ministry of Education Key Laboratory of Big Data and Intelligent Robot and the Guangdong Provincial Key Lab of Computational Intelligence and Cyberspace Information.}
\thanks{ Shengfeng He is with the Singapore Management University, Singapore. E-mail: shengfenghe@smu.edu.sg.} 
% the School of Computing and Information Systems, 
}

\markboth{IEEE Transactions on Pattern Analysis and Machine Intelligence}%
{Shell \MakeLowercase{\textit{Liu et al.}}: Rotation-Adaptive Point Cloud Domain Generalization via Intricate Orientation Learning}

\IEEEtitleabstractindextext{
\begin{abstract}
  The vulnerability of 3D point cloud analysis to unpredictable rotations poses an open yet challenging problem: \emph{orientation-aware 3D domain generalization}. Cross-domain robustness and adaptability of 3D representations are crucial but not easily achieved through rotation augmentation. Motivated by the inherent advantages of intricate orientations in enhancing generalizability, we propose an innovative rotation-adaptive domain generalization framework for 3D point cloud analysis. Our approach aims to alleviate orientational shifts by leveraging intricate samples in an iterative learning process. Specifically, we identify the most challenging rotation for each point cloud and construct an intricate orientation set by optimizing intricate orientations. Subsequently, we employ an orientation-aware contrastive learning framework that incorporates an orientation consistency loss and a margin separation loss, enabling effective learning of categorically discriminative and generalizable features with rotation consistency. Extensive experiments and ablations conducted on 3D cross-domain benchmarks firmly establish the state-of-the-art performance of our proposed approach in the context of orientation-aware 3D domain generalization. 
\end{abstract}

\begin{IEEEkeywords}
  Point cloud domain generalization, contrastive learning, rotation robustness, intricate orientation mining
\end{IEEEkeywords}
}

\maketitle

\IEEEdisplaynontitleabstractindextext

\IEEEpeerreviewmaketitle

\section{Introduction}

\indent3D scene understanding and reasoning are crucial in various applications~\cite{leibe2007dynamic, brostow2008segmentation} such as autonomous driving, architectural design, and virtual/augmented reality. 3D Point cloud representation has gained popularity due to its simplicity and effectiveness, particularly with the advancements in deep learning~\cite{wang2019dynamic, qi2017pointnet, zhang2022self}. However, most existing 3D point cloud analysis models hold a strong \textit{i.i.d.} (independent and identically distributed) assumption between training and testing domains. This assumption can lead to significant performance degradation in real-life out-of-distribution scenarios. Factors such as varying sensor parameters and environmental conditions introduce distributional divergence between training and testing domains, limiting the practical usage of these models.

\begin{figure}[t]
    % \flushleft
    \centering
    \subfloat[Trained on aligned data]{%[b]{0.45\textwidth}
        \label{fig:teaser_a}
        % \centering
        \includegraphics[width=0.55\linewidth]{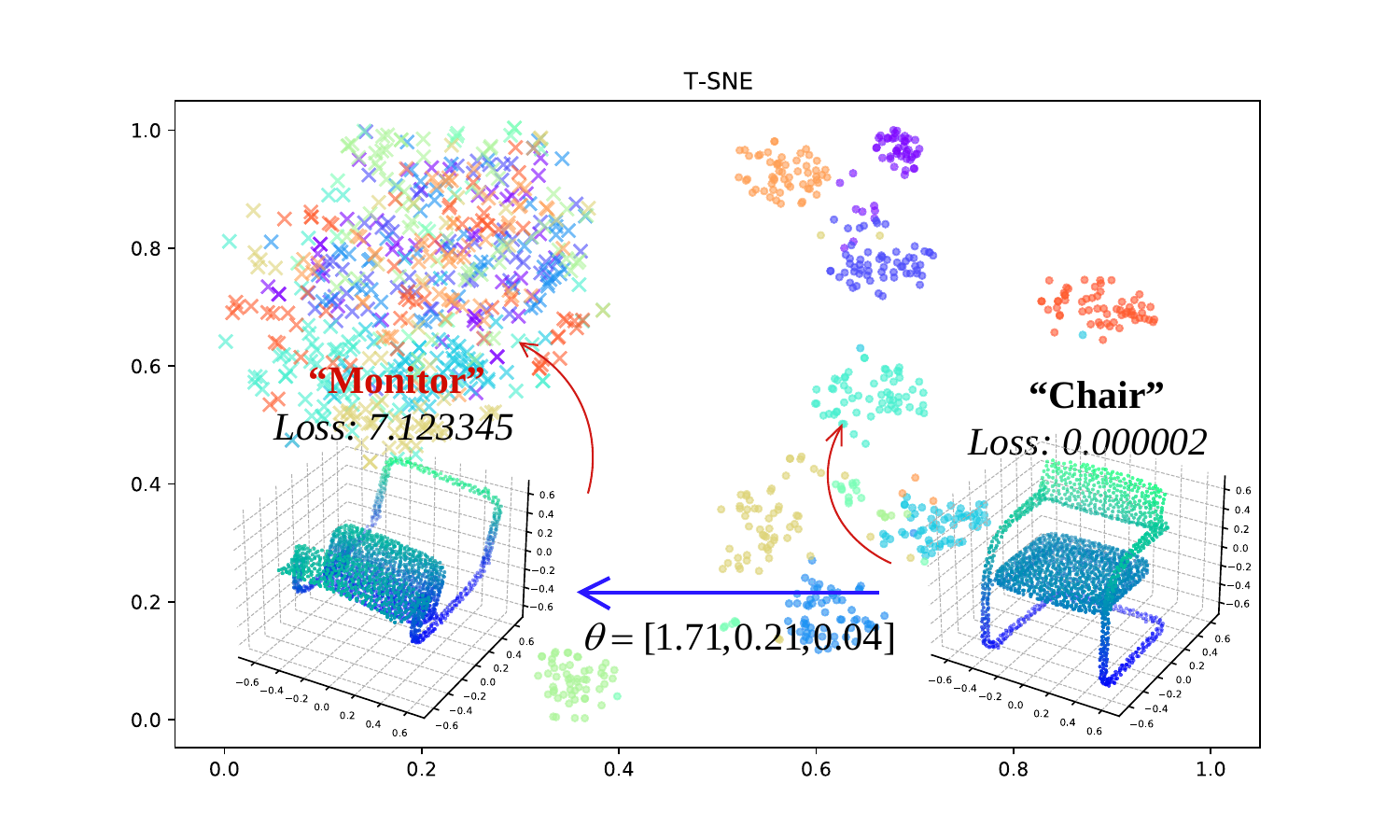}
    }
    \subfloat[w/ rotation augmentation]{%[b]{0.45\textwidth}
        \label{fig:teaser_b}
        % \centering
        \includegraphics[width=0.42\linewidth]{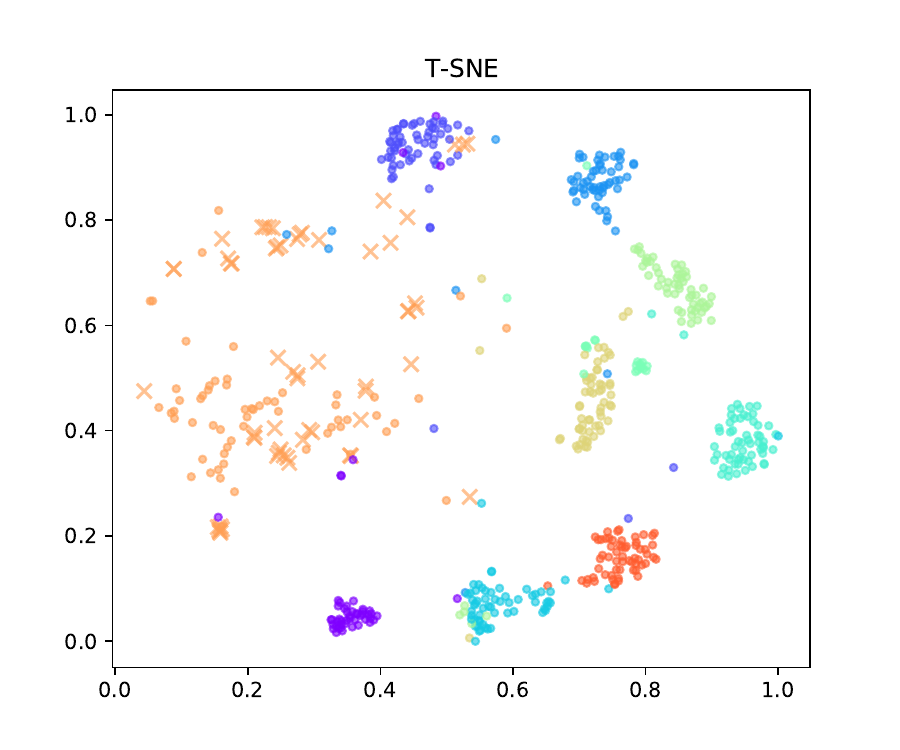}
        \hspace{-20mm}

        \includegraphics[width=0.2\linewidth]{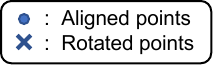}

    }
    \vspace{-2mm}
    \caption{t-SNE visualization of the feature spaces in the geometric encoder of DGCNN~\cite{wang2019dynamic} trained on ModelNet~\cite{wu20153d} before and after random rotation.
(a) Random rotation perturbation leads to a chaotic cluster in the feature space, rendering it non-discriminative.
(b) Even with rotation augmentation, the rotated shapes tend to be located in the low-density region of the feature space, which remains non-discriminative and less consistent. }
\vspace{-5mm}
    \label{fig:teaser}
\end{figure}

Prevailing methods focus on addressing different types of domain shifts in point clouds, such as shape variance, non-uniform point density, sensor noise, and self-occlusion, through point cloud domain adaptation (3DDA)~\cite{qin2019pointdan, liang2022point} and domain generalization (3DDG)~\cite{huang2021metasets, wei2022learning}. 
However, they make the naive assumption that objects from both source and target domains are strictly aligned, meaning they have the same upright pose across all objects and categories. This assumption disregards the orientation of objects and can result in a model that is biased toward the aligned dataset. As a consequence, even slight variations in orientation may incur catastrophic misclassification by the model (Fig.~\ref{fig:teaser}(a)). In real-world scenarios, objects often originate from various domains with diverse orientations, ensuring an accurate understanding of these targets imposes great demands on a generalizable and rotational robust 3D recognition systems. This motivates us to investigate the orientational gap and study a new problem called \textit{orientation-aware 3D domain generalization}, which aims at exploring the robustness of 3D representations under the disturbance of varying rotations.

To ensure the robustness of a model against arbitrary rotational variations, random rotation augmentation~\cite{qi2017pointnet,wang2019dynamic} is intuitively applied. However, generalizing to arbitrary rotations is challenging due to the numerous possible rotated angles in 3D space~\cite{zhao2019rotation}. Sparse sampling over the dense rotation space often leads to imbalanced learning (Fig.~\ref{fig:teaser}(b)), where the model tends to memorize the easy tasks but struggles with unseen challenges. In practice, tackling challenging tasks can enhance one's ability to draw inferences for similar problems, a process known as extrapolation. Deep models exhibit a similar talent for extracting rich information from samples that are difficult to separate~\cite{schroff2015facenet}. This motivates our focus on challenging orientations, which will be discussed with supporting experimental evidence in Section~\ref{sec:analysis}. Building upon these observations, we aim to enhance the model's generalizability to arbitrary rotations by learning from intricate samples.

Based on this idea, we propose to address the problem of orientation-aware 3D domain generalization through intricate orientation mining. Our approach alternatively optimizes between multiple steps. First, we identify the most aggressive rotation for each point cloud and construct an intricate set by optimizing intricate orientations. Next, we create the hardest sample pairs based on the intricate set and utilize contrastive learning to obtain categorically discriminative and generalizable representations with rotation consistency.
More specifically, we optimize the intricate rotation matrix, which is parameterized by three optimizable parameters, by maximizing the prediction error of the current model. Once we obtain the intricate set, we construct sample-wise and category-wise pairs for contrastive training. The sample-wise pair consists of the original point cloud along with its rotated variant, where the rotation angles are sampled from the intricate set. The category-wise pair is formed by selecting point clouds from different classes that are close together in the representation space.

In summary, our contributions are three folds:
\begin{itemize}
\item We introduce the novel task of orientation-aware 3D domain generalization, which more accurately captures real-world domain shifts involving unpredictable rotations, and delve into the importance of challenging orientations in reducing orientational shift.
\item We propose a novel approach that enhances rotational robustness through intricate orientation mining, which further incorporates a contrastive framework that utilizes intricate samples to improve the generalizability of feature learning via inner- and intra-categorically contrastive learning.
\item Experimental results on widely used benchmarks demonstrate that our method achieves state-of-the-art performance on several point cloud analysis tasks under the orientation-aware 3D domain generalization setting.
\end{itemize}

\section{Related Works}

\noindent\textbf{3D Point Cloud Domain Adaptation and Generalization.}
Early endeavors within 3D domain adaptation (3DDA) focused on extending 2D adversarial methodologies~\cite{qin2019pointdan} to align point cloud features. Alternative methods have delved into geometry-aware self-supervised pre-tasks. Achituve \etal~\cite{achituve2021self} introduced DefRec, a technique employing self-complement tasks by reconstructing point clouds from a non-rigid distorted version, while Zou \etal~\cite{zou2021geometry} incorporating norm curves prediction as an auxiliary task. Liang \etal~\cite{liang2022point} put forth MLSP, focusing on point estimation tasks like cardinality, position, and normal. SDDA~\cite{cardace2023self} employs self-distillation to learn the point-based features. Additionally, post-hoc self-paced training~\cite{zou2021geometry,fan2022self,park2023pcadapter} has been embraced to refine adaptation to target distributions by accessing target data and conducting further finetuning based on prior knowledge from the source domain.
In contrast, the landscape of 3D domain generalization (3DDG) research remains nascent. Metasets~\cite{huang2021metasets} leverage meta-learning to address geometric variations, while PDG~\cite{wei2022learning} decomposes 3D shapes into part-based features to enhance generalization capabilities.
Despite the remarkable progress, existing studies assume that objects in both the source and target domains share the same orientation, limiting their practical application. This limitation propels our exploration into orientation-aware 3D domain generalization through intricate orientation learning.

\noindent\textbf{Rotation-generalizable Point Cloud Analysis.}
Previous works in point cloud analysis~\cite{qi2017pointnet, wang2019dynamic} enhance rotation robustness by introducing random rotations to augment point clouds. {However, generating a comprehensive set of rotated data is impractical, resulting in variable model performance across different scenes. To robustify the networks \wrt randomly rotated point clouds,} rotation-equivariance methods explore equivalent model architectures by incorporating equivalent operations~\cite{su2022svnet, Deng_2021_ICCV, luo2022equivariant} or steerable convolutions~\cite{chen2021equivariant, poulenard2021functional}.
Alternatively, rotation-invariance approaches aim to identify geometric descriptors invariant to rotations, such as distances and angles between local points~\cite{chen2019clusternet, zhang2020global} or point norms~\cite{zhao2019rotation, li2021rotation}. Besides, {Li \etal~\cite{li2021closer} have explored disambiguating the number of PCA-based canonical poses, while Kim \etal~\cite{kim2020rotation} and Chen \etal~\cite{chen2022devil} have transformed local point coordinates according to local reference frames to maintain rotation invariance. However, these methods focus on improving in-domain rotation robustness, neglecting domain shift and consequently exhibiting limited performance when applied to diverse domains. This study addresses the challenge of cross-domain generalizability together with rotation robustness and proposes novel solutions.} 

\noindent\textbf{Intricate Sample Mining}, aimed at identifying or synthesizing challenging samples that are difficult to classify correctly, seeks to rectify the imbalance between positive and negative samples for enhancing a model's discriminability. While traditional works have explored this concept in SVM optimization~\cite{felzenszwalb2009object}, shallow neural networks~\cite{dollar2009integral}, and boosted decision trees~\cite{yu2019unsupervised}, recent advances in deep learning have catalyzed a proliferation of researches in this area across various computer vision tasks. For instance, 
Lin \etal~\cite{lin2017focal} proposed a focal loss to concentrate training efforts on a selected group of hard examples in object detection, while Yu \etal~\cite{yu2019unsupervised} devised a soft multilabel-guided hard negative mining method to learn discriminative embeddings for person Re-ID. Schroff \etal~\cite{schroff2015facenet} introduced an online negative exemplar mining process to encourage spherical clusters in face embeddings for individual recognition, and Wang \etal~\cite{wang2021instance} designed an adversarially trained negative generator to yield instance-wise negative samples, bolstering the learning of unpaired image-to-image translation. In contrast to existing studies, our work presents the first attempt to mitigate the orientational shift in 3D point cloud domain generalization, by developing an effective intricate orientation mining strategy to achieve orientation-aware learning.

\section{Orientational Shift Analysis} \label{sec:analysis}

\textbf{Problem Definition.}
In practical scenarios, objects often originate from various domains, and concurrently, present diverse orientations. To achieve an accurate understanding of these targets, our goal is to grant 3D recognition systems with cross-domain generalizability and robustness to rotational transformations. To this end, we introduce a novel task: orientation-aware 3D domain generalization, and explore the applicability of existing 3D recognition systems under orientational shifts.
Let $\mathbb{X}$ and $\mathbb{Y}$ be the input and label spaces, respectively. We consider a labeled source domain $D_s = {\{(P_i^s, y_i^s)}\}^{n_s}_{i=1}$ with $n_s$ samples and an unlabeled target domain $D_t = \{{P_i^t}\}^{n_t}_{i=1}$. Here, $y^s\in \mathbb{Y}$ is the source labels, and $P^s, P^t \in \mathbb{X}$ represent point clouds with a specified orientation relative to the world coordinate system, where each $P_i \in \mathbb{R}^{N_c \times 3}$ consists of $N_c$ points. The orientation is defined by a $3\times3$ rotation matrix $M_i \in \mathcal{O} \subseteq$ SO(3), where $\mathcal{O}$ is the set of orientations in the given domain and SO(3) represents all rotations in $\mathbb{R}^3$. The objective of orientation-aware 3D domain generalization is to learn a projection function $g: \mathbb{X} \to \mathbb{Y}$ that can be applied to any given target domain $D_t$ with arbitrary orientations in $\mathcal{O}$, by solely training on the labeled source domain $D_s$.

\begin{figure}[t]
    % \flushleft
    \centering
    \subfloat[Random rotation augmentation]{%[b]{0.45\textwidth}
        \label{fig:statement_random}
        % \centering
        \includegraphics[width=0.23\textwidth]{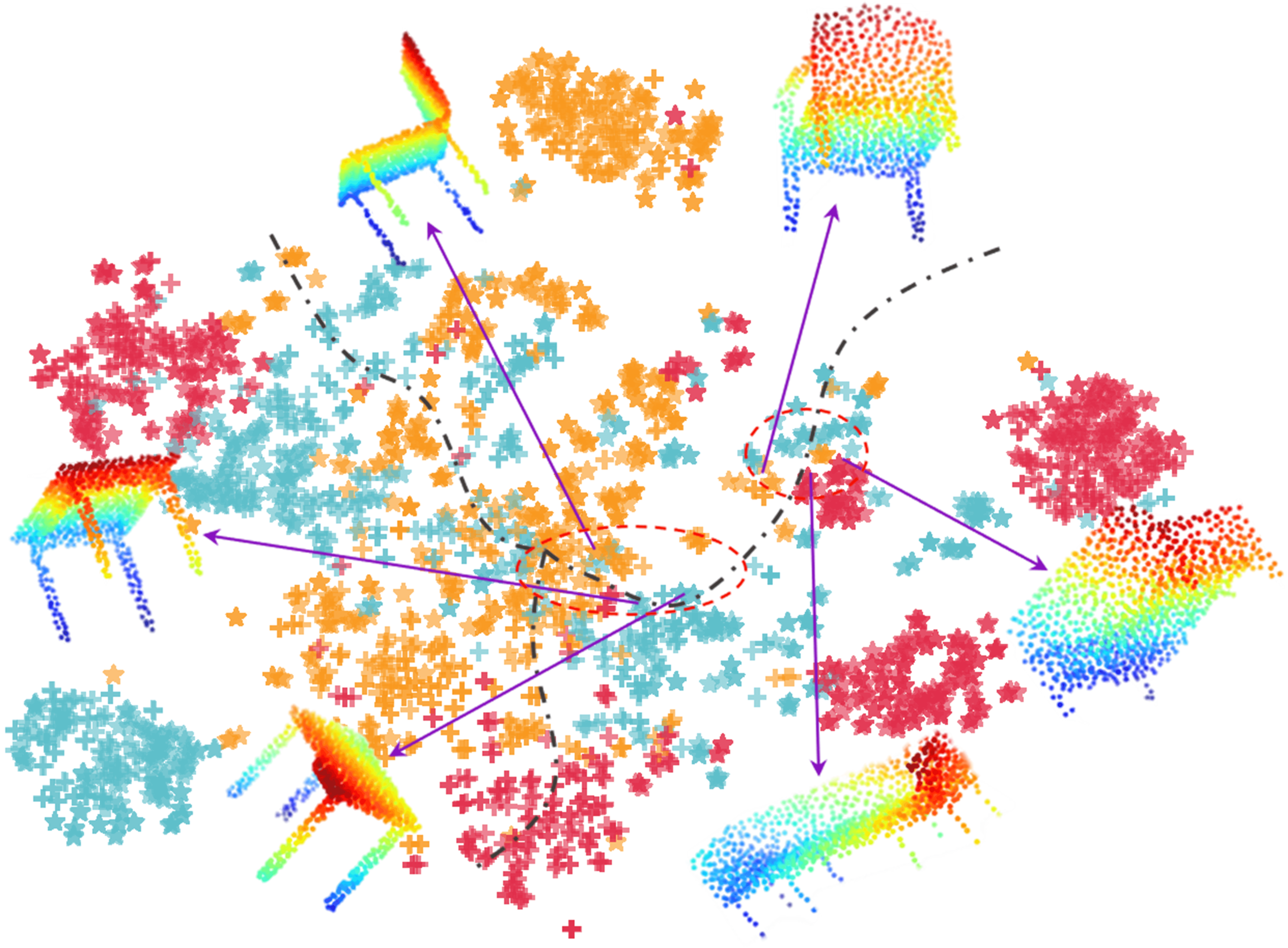}
    }
    % \hspace{5mm}
    \subfloat[Intricate orientation training]{%[b]{0.45\textwidth}
        \label{fig:statement_ours}
        % \centering
        \includegraphics[width=0.23\textwidth]{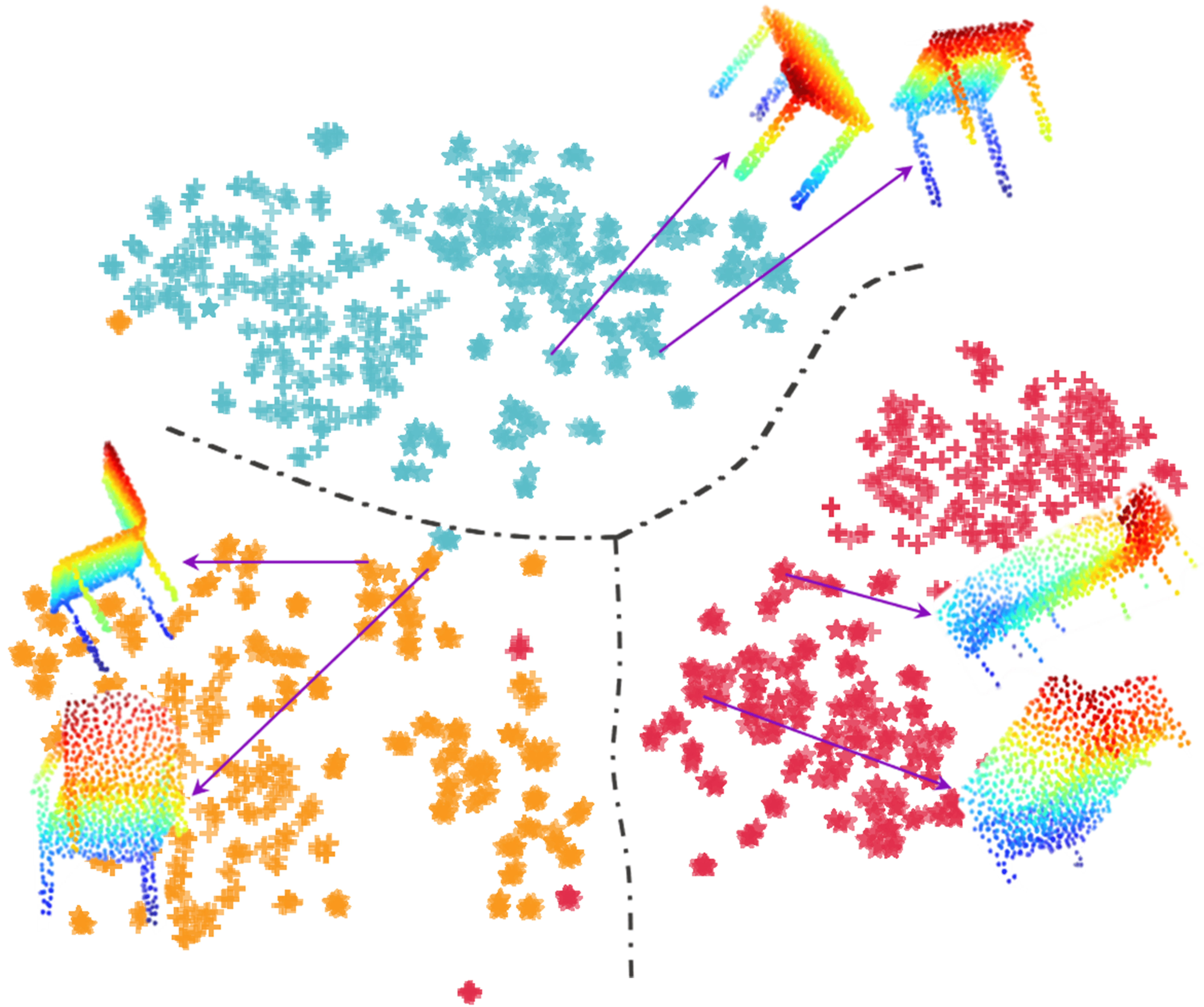}
    }
    \vspace{-2mm}
    \caption{t-SNE visualization of the feature spaces trained with (a) random rotation augmentation and (b) our proposed intricate orientation training. For the sake of simplicity, we only visualize samples from three categories (table, chair, and sofa), which share part of distinguishing features (\eg, slim legs, plane seats, \etc). Samples from ModelNet (the source domain) and ShapeNet (the target domain) are denoted by star (`$\star$') and plus (`$+$'), respectively. The gray dash line denotes the approximated decision boundaries.}
    \label{fig:feature_statement}
    \vspace{-5mm}
\end{figure}

\noindent\textbf{Generalization Analysis of Orientations.}
To accurately understand 3D shapes, information from multiple perspectives is often necessary. Some of the perspectives are easy-to-understand by deep models but less informative, incurring a phenomenon known as ``taking the whole from a part''. Specifically, in the context of rotation-robust point cloud analysis, these learning perspectives refer to those rotated variants with small gradients during training. Random rotation attempts to capture comprehensive information through a Monte Carlo approach, but the vast number of possible rotated angles introduces difficulties and leads to imbalanced learning.
This biased learning problem could give rise to inaccurate selection of point features, making false decisions. As shown in Fig.~\ref{fig:feature_statement}(a), the learned model cannot well distinguish the rotated sample from ``table" and ``chair", as they may contain similar features such as slim legs, which are learned from easy-to-understand samples. 
On the other hand, intricate samples lying beyond current cognitive boundaries can offer complementary knowledge for learning a more accurate discriminative boundary.

\begin{figure}[t]
    \captionsetup[subfloat]{justification=centering,singlelinecheck=false}
    % \flushleft
    \centering
    \subfloat[Per-class maximum mean discrepancy]{%[b]{0.45\textwidth}
        \label{fig:analysis_a}
        % \centering
        \includegraphics[width=0.32\textwidth]{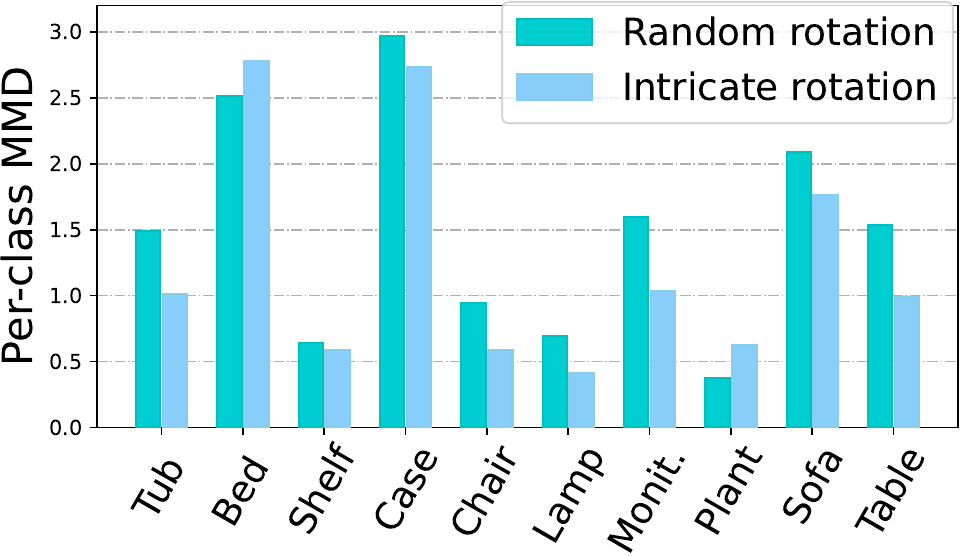}
    }
    % \hspace{.4in}
    \subfloat[Consistency \\\& discrimination]{%[b]{0.45\textwidth}
        \label{fig:analysis_b}
        % \centering
        \includegraphics[width=0.16\textwidth]{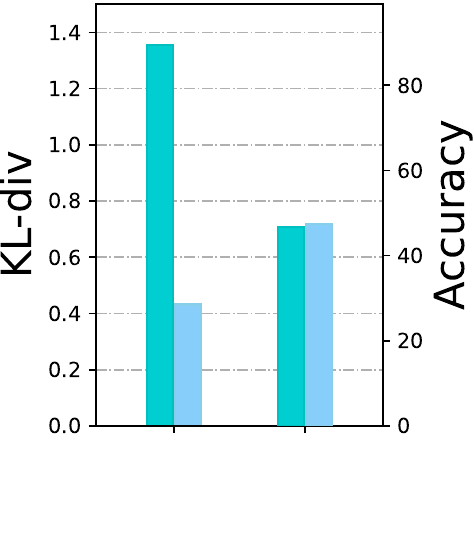}
    }
    \vspace{-3mm}
    \caption{(a) Maximum Mean Discrepancy~\cite{borgwardt2006integrating} between the features of the ModelNet and ShapeNet subdomains in PointDA~\cite{qin2019pointdan}, learned with different orientation augmentations. (b) The mean consistency rate was computed using KL-divergence between multiple augmented variants and the mean accuracy on ShapeNet.}
    \vspace{-5mm}
    \label{fig:analysis}
\end{figure}

We empirically validate these insights by employing Maximum Mean Discrepancy (MMD)~\cite{gretton2012kernel} as a measure of distributional shifts. The MMD calculates the distance between source and target distributions in the Reproducing Kernel Hilbert Space (RKHS). We train a DGCNN~\cite{wang2019dynamic} on the ModelNet (source domain) and test it on ShapeNet (target domain), augmenting the data with random rotations and intricate orientations, respectively. The intricate orientations of the training dataset are obtained by optimizing the rotational angles of each point cloud to maximize the cross-entropy loss with a baseline model trained on aligned data. Fig.~\ref{fig:analysis}(a) depicts the per-class MMD values, demonstrating that training with intricate orientations reduces orientational shifts more effectively. To evaluate discriminability and consistency under varying rotations, we further train a linear SVM on the source domain and test it on the target domain. We augment each point cloud 64 times as illustrated in Section~\ref{sec:experiment} and measure consistency by calculating the mean KL-divergence between their output probabilities. As shown in Fig.~\ref{fig:analysis}(a), the discriminability of the model augmented with intricate orientations matches that of the randomly rotated version, while better preserving consistency across different rotations.
Based on these findings, we aim to leverage intricate samples to reduce orientational shift, enhance output consistency, improve model discriminability to arbitrary rotations, and finally obtain a better classifier that is generalizable towards various domains (Fig.~\ref{fig:analysis}(b)).

% \vspace{-3mm}
\section{Intricate Orientation Learning} \label{sec:method}

\begin{figure*}[t]
    % \flushleft
    \centering
    \includegraphics[width=1\textwidth]{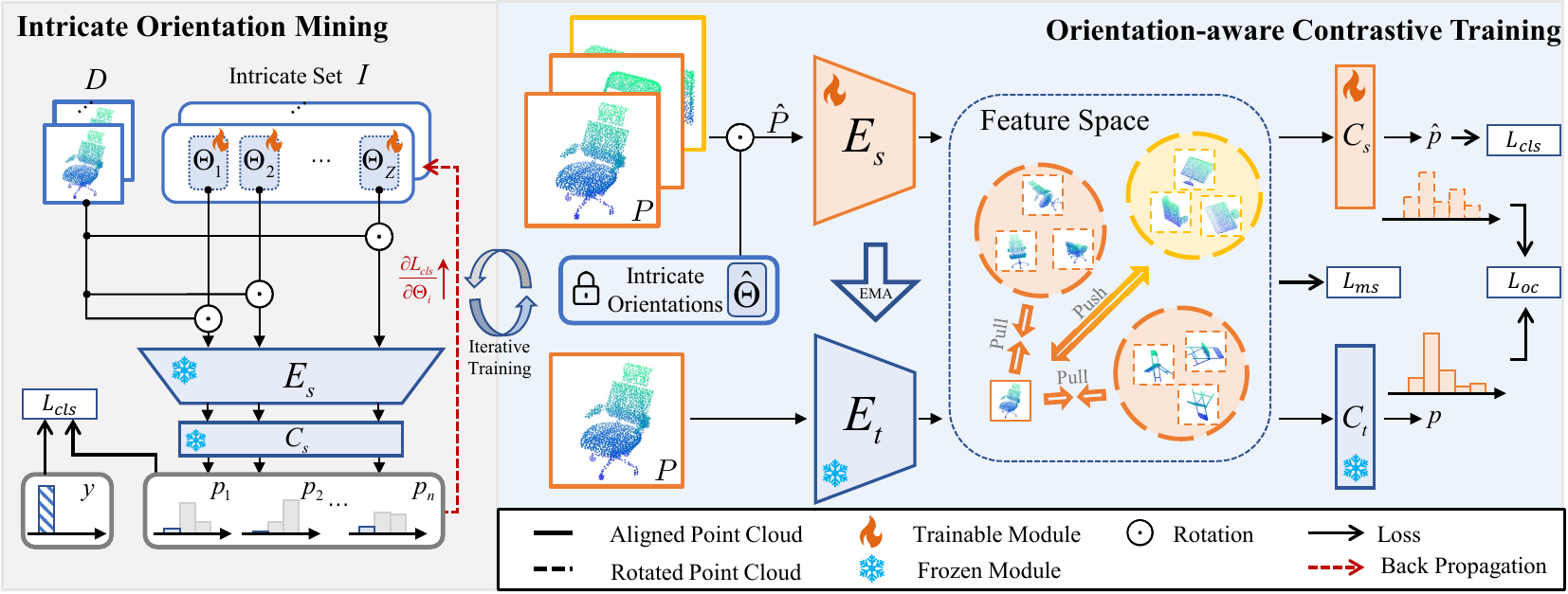}
    \vspace{-6mm}
    \caption{Pipeline of our intricate orientation learning framework for point cloud classification, which alternatively optimizes between the intricate orientation mining (Left) and orientation-aware contrastive training (Right). }
    \label{fig:pipeline}
    \vspace{-6mm}
\end{figure*}

The pipeline of our framework is depicted in Fig.~\ref{fig:pipeline}. It consists of an iterative optimization process that alternates between intricate orientation mining and orientation-aware contrastive training. We first generate an alternative set of intricate rotation angles through intricate orientation mining. Then we augment the training point clouds by applying the selected angles from the alternative set, creating a series of intricate sample pairs. To enhance the generalizability of the learned representations, we employ orientation-aware contrastive training using these intricate pairs. Our framework incorporates an orientation consistency loss, which encourages the learning of more representative features while promoting their consistency \wrt random rotations. Additionally, we introduce a margin separation loss to further improve the categorical discriminability of the learned representation space.
In the following sections, we elaborate on each component of our framework in detail.

% \vspace{-3mm}
\subsection{Intricate Orientation Mining}
Let us first consider a standard classification problem over the given data distribution $D = {\{(P_i, y_i)}\}^{n}_{i=1}$. Suppose we have a point cloud classification model, the goal is to search the model parameters $\omega^*$ that optimize the empirical risk $E_{(P, y)\sim D}\left[L(w, P, y)\right]$, where $L$ is a proper loss function. To improve the robustness, the most adopted technique is data augmentation, such that the model can resist possible perturbations on the input data, resulting in the following objective:
\begin{equation}
    \omega^*=\mathop{\arg\min_{\omega}} \mathbb{E}_{(P, y)\sim D} \left[L(\omega, f(P), y)\right],
\end{equation}
where $f$ is the perturbation function. Considering the orientational shift only, the perturbation function of a given point cloud $P_i$ can be regarded as augmenting the aligned pose with a rotation matrix $M_i$. According to Euler's rotation theorem, the rotation matrix is defined in $\mathbb{R}^3$ as $M_i = R_{\theta_{x_i}}\cdot R_{\theta_{y_i}}\cdot R_{\theta_{z_i}}$, 
where $R_{\theta_j}$ is the rotation matrix about the axis-j by angle $\theta_{j} \in \left[-\pi, \pi\right)$ over the cartesian coordinates. In this case, the rotation matrix is parameterized by $\Theta_i=\left[\theta_{x_i}, \theta_{y_i}, \theta_{z_i}\right]$. Since the trigonometric function is differential in the scope $\left[-\pi, \pi\right)$, we can optimize $\Theta_i$ through gradient descent.

The essence of intricate orientation mining lies in the deliberate search for intricate orientations, instead of directly optimizing the model with random rotations. This endeavor aims to identify a rotation matrix $\hat{\Theta}_i$ that introduces perturbations to the point cloud, such that the 3D model is confounded:
\begin{equation}
    \hat{\Theta}_i = \mathop{\arg\max_{\Theta_i}}L(\omega^*, f(\Theta_i, P_i), y_i), \label{formula:iom}
\end{equation}
where $f(\Theta_i, P_i)=M_i\cdot P_i$, and $L$ is a task-specific optimization function~(\eg, cross-entropy loss for classification task, hereafter referred to as $L_{cls}$). By iteratively maximizing this objective, we can calculate the most intricate orientations for all the point clouds within $D$. The details of gradient calculation are provided in the supplement. We adopt Projected Gradient Descent~\cite{madrytowards} to ensure that the optimized $\Theta_i$ can generate a pure rotation matrix. By applying intricate orientation mining to the whole training dataset, we obtain the intricate orientation set $I=\{\hat{\Theta}_{i}\}^{n}_{i=1}$, which is essential for the construction of intricate sample pairs. Note that each orientation is specified for the according sample. To increase the diversity of $I$, we further initialize $\Theta$ with random values and repeat the optimization multiple times. Finally, the intricate set is defined as $I=\{\{\hat{\Theta}^j_{i}\}^{AT}_{j=1}\}^{n}_{i=1}$, where $AT$ is the number of augmentation times. During the training stage, we periodically update the intricate set $I$ for every $T$ epochs.

% \vspace{-3mm}
\subsection{Orientation-aware Contrastive Training}

Based on the intricate orientation set $I$, we construct the intricate sample pairs and build a contrastive learning framework upon them to obtain categorically discriminative and generalizable features with rotational consistency. As Fig.~\ref{fig:pipeline} shows, our model consists of an optimizable student network $F_s\!=\!C_s\circ E_s$ and a frozen teacher network $F_t\!=\!C_t\circ E_t$, where $E_{s/t}$ is a feature extractor which encodes point cloud to the representation space and derives the final classification result through a linear classifier $C_{s/t}$. The teacher network is optimized by Exponential Moving Average (EMA)~\cite{tarvainen2017mean}.

\noindent\textbf{Inner-sample Rotational Consistency.} The main idea is to maximize the agreement between the original shape and its intricate augmented variants via self-contrastive learning. Each time we sample a mini-batch of $N$ point clouds $\left\{P_{m}\right\}^N_{m=1}$ and randomly select one intricate rotation $\hat{\Theta}_m$ from $I$ for each point cloud. Then we construct the intricate sample pairs $B_{it} = \{(P_{m}, \hat{P}_m )\}^N_{m=1}$, where $\hat{P}_m = f(\hat{\Theta}_{m}, P_m)$ is the augmented version of the original point cloud with the intricate rotation. $B_{it}$ serve as the input of (${F_t}$, $F_s$) and the output logits of $(P_{m}, \hat{P}_m )$ are denoted as $p_m$ and $\hat{p}_m$, respectively. To ensure the consistency of the intricate pairs, we perform knowledge distillation~\cite{hinton2015distilling} on the output probabilities, the orientation consistency loss is therefore formulated as:
\begin{equation}\label{eq5-0}
    \sigma(p/\tau) = \frac{exp(p/\tau)}{\sum_{j=1}^{K}{exp(p^{(j)}/\tau)}},
\end{equation}
\begin{equation}
    L_{oc} = -\frac{1}{N}\sum_{m=1}^{N} \sigma(\hat{p}_m/\tau_t)\log(\sigma(p_m /\tau_s)), \label{eq5}
\end{equation}
where $K$ is the number of categories and $\sigma(\cdot)$ is the softmax function. $\tau \in \mathbb{R^+}$ is the temperature parameter controlling the magnitude of the output logits, where the subscripts in $\left[\tau_s, \tau_t\right]$ denote the source and target domains, respectively. By iteratively refining the distance between the aligned object and its intricate variant, the model can gradually adapt to the harder rotational variations and learn common knowledge concerning different rotations.

\noindent\textbf{Intra-category Discriminability.} To further enhance discriminability, we aim for the learned representation space to maintain rotational consistency while improving the classification ability. To achieve this, we introduce a margin separation loss that yields a classifier with a concise and accurate boundary, capable of handling the unforeseen rotational perturbations. Specifically, we first construct the intra-category intricate pairs by considering the relationships between different categories in the same mini-batch. The intricate pairs can be divided into two groups, including the positive pairs and the negative pairs.
Simply, the positive pairs can be found by traversing the training batch and sorting out samples that belong to the same class. But the embedding space under multiple orientations may not be well simulated by the originally aligned point clouds. Instead of learning from only one intricate orientation, we further augment each point cloud $P_{i}$ with $V$ number of intricate angles sampled from $\{\hat{\Theta}^l_{i}\}^{AT}_{l=1}$ to better simulate the feature space. The loss is designed to minimize the cosine distance between the positive samples as follows:
\begin{equation}
    L_{pos}\!=\!- \frac{1}{K}\sum_{k=1}^{K} \!\frac{1}{V{N^k_p}^2}\!\!\!\sum_{i=1,j=1 \atop j\neq i}^{N^k_{p}, N^k_{p}}\!\sum_{v=1}^{V} \cos(\frac{E_s(\hat{P}_{i,v})}{\tau'},\! \frac{E_s(\hat{P}_j)}{\tau'}),
    \label{eq6}
\end{equation}
where $\cos(\cdot, \cdot)$ denotes the cosine similarity between the two inputs, $N^k_{p}$ is the number of samples of class $k$ within the mini-batch and $\tau'$ is the temperature parameter.

The negative pairs are formed by samples with different labels.
To enlarge the discrepancy of different categories in the representation space, we plan to maximize the distance between each negative pair with:
\begin{equation}
    L_{neg}\!=\!\frac{1}{K}\sum_{k=1}^{K}  \frac{1}{{N^k_p}{N^k_n}}\sum_{i=1}^{N^k_p} \sum_{j=1}^{N^k_{n}} \cos( \frac{E_s(\hat{P}_i)}{\tau'}, \frac{E_s(\hat{P}_j)}{\tau'}),
    \label{eq7}
\end{equation}
where $N^k_{n}$ is the number of negative samples.
The overall margin separation loss is summarized as follows:
\begin{equation}
    L_{ms} = L_{pos} + L_{neg}.
\end{equation}Through contrastive tasks, the point representations under varying rotations are further clustered within the same category, while the other categories are pushed away. A more compact and consistent representation space is formed, thereby enhancing the discriminability for rotated shape classification.

\begin{table*}[t] % table for OSDA setting on Office31
    \caption{Comparison of the macro-average precision score \textit{Avg.}~($\%$) under the orientation-aware 3D domain generalization setting. The value after $\pm$ denotes the standard deviation on the 64 evaluated series. RE, RI, DA and DG represent Rotation-Equivalent, Rotation-Invariant, 3D Domain Adaptation, and 3D Domain Generalization. The top 2 records are marked in \bc{red} and \rc{blue}, respectively.}
    \label{tab:pointda10_avg} 
    \vspace{-4mm}
    \small
    \begin{center}
    \setlength{\tabcolsep}{0.4cm}{
    \resizebox{1\textwidth}{!}{
    \begin{tabular}{l|c|l|l|l|l|l|l|c}
    \hline
    \multicolumn{9}{c}{DGCNN~\cite{wang2019dynamic}}\\
    \hline
    \makecell[c]{Methods}
    &Type & \makecell[c]{M$\to$S} & \makecell[c]{M$\to$S*} & \makecell[c]{S$\to$M} & \makecell[c]{S$\to$S*} & \makecell[c]{S*$\to$M} & \makecell[c]{S*$\to$S} & {AVG} \\

    \hline
    Supervised                                &\multirow{2}{*}{-}                  &{74.4 $\pm$ 5.5}   &{56.7 $\pm$ 3.9}   &{87.3 $\pm$ 14.5}   &{56.7 $\pm$ 3.9}  &{87.3 $\pm$ 14.5}  &{74.4 $\pm$ 5.5}    &{72.8}\\
    w/o Adapt                                 &                  &{54.7 $\pm$ 9.2}   &{25.3 $\pm$ 2.3}   &{57.6 $\pm$ 8.6}   &{27.9 $\pm$ 2.8}  &{45.3 $\pm$ 4.4}  &{37.1 $\pm$ 2.8}    &{41.3}    \\
    \hline
    VN~\cite{Deng_2021_ICCV}                  &\multirow{3}{*}{RE}                  &{61.2 $\pm$ 0.0}    &{27.3 $\pm$ 0.0} &{68.9 $\pm$ 0.0} &{23.5 $\pm$ 0.0} &{35.1 $\pm$ 0.0} &{31.5 $\pm$ 0.0} &{41.3}\\
    SVN~\cite{su2022svnet}                  &                  &{58.4 $\pm$ 0.7}    &{26.3 $\pm$ 0.6} &{64.9 $\pm$ 0.8}  &{23.6 $\pm$ 0.4} &{33.1 $\pm$ 0.8}    &{31.1 $\pm$ 0.8} &{39.6}\\
    EOMP~\cite{luo2022equivariant}            &                  &{56.7 $\pm$ 0.6}    &{27.0 $\pm$ 0.5} &{64.2 $\pm$ 0.9}  &{26.6 $\pm$ 0.5} &{25.6 $\pm$ 0.6}    &{26.9 $\pm$ 0.4} &{37.8}\\
    \hline
    SPRIN~\cite{you2021prin}              &\multirow{5}{*}{RI}             &{58.6 $\pm$ 0.7} &{27.3 $\pm$ 0.7}   &{73.9 $\pm$ 0.6} &{20.7 $\pm$ 0.9}    &{45.7 $\pm$ 0.7} &{37.3 $\pm$ 0.8}  &{43.9}\\
    RIPCA~\cite{li2021closer}              &            &{62.5 $\pm$ 1.2} &\rc{30.5 $\pm$ 1.2}   &{72.9 $\pm$ 0.8} &{27.2 $\pm$ 1.1}    &{47.5 $\pm$ 1.8} &{39.0 $\pm$ 1.2}  &\rc{46.6}\\
    RIConv++~\cite{zhang2022riconv}        &             &{38.9 $\pm$ 0.4} &{19.0 $\pm$ 0.6}   &{57.2 $\pm$ 0.8} &\rc{30.2 $\pm$ 1.0}    &{32.5 $\pm$ 0.7} &{38.4 $\pm$ 0.7}  &{36.0}\\ % special design for Pointnet++
    PaRI~\cite{chen2022devil}              &             &{40.9 $\pm$ 0.0} &{25.0 $\pm$ 0.5}   &{55.8 $\pm$ 0.7} &\bc{32.5 $\pm$ 0.7}    &{44.0 $\pm$ 1.0} &{39.4 $\pm$ 0.1}  &{39.6}\\
    LocoTrans~\cite{chen2024local}              &                  &\rc{65.2 $\pm$ 0.2}   &{30.4 $\pm$ 0.5}    &\rc{76.5 $\pm$ 0.4}  &{25.7 $\pm$ 0.4}  &{37.7 $\pm$ 0.7} &{26.9 $\pm$ 0.0}    &{43.7}\\ 
    \hline
    PointDAN~\cite{qin2019pointdan}           &\multirow{6}{*}{DA}                   &{50.5 $\pm$ 9.4}   &{27.5 $\pm$ 2.5}   &{58.9 $\pm$ 7.0}   &{26.1 $\pm$ 2.4}   &{36.6 $\pm$ 5.1}   &{34.7 $\pm$ 3.1}    &{39.1}    \\
    DefRec~\cite{achituve2021self}            &                   &{53.3 $\pm$ 9.2}   &{24.0 $\pm$ 3.6}   &{58.2 $\pm$ 8.0}   &{21.5 $\pm$ 2.9}   &{37.4 $\pm$ 6.3}   &{30.0 $\pm$ 3.0}    &{37.4}     \\
    GAST~\cite{zou2021geometry}               &                   &{37.9 $\pm$ 6.3}   &{20.7 $\pm$ 1.9}   &{47.4 $\pm$ 2.5}   &{14.7 $\pm$ 2.0}   &{28.9 $\pm$ 1.1}   &{26.0 $\pm$ 1.0}    &{29.3}     \\
    MLSP~\cite{liang2022point}                &                   &{60.9 $\pm$ 9.2}   &{30.0 $\pm$ 3.9}   &{62.6 $\pm$ 5.3}   &{23.4 $\pm$ 3.7}   &{44.9 $\pm$ 4.7}   &{37.3 $\pm$ 3.1}    &{43.2}     \\
    SDDA~\cite{cardace2023self}               &                   &{58.4 $\pm$ 9.8}   &\rc{30.5 $\pm$ 3.8}   &{64.4 $\pm$ 5.3}   &{25.9 $\pm$ 3.8}   &{40.1 $\pm$ 6.1}   &{39.5 $\pm$ 3.4}    &{43.1}    \\
    PCFEA~\cite{wang2024progressive}     &                   &{62.2 $\pm$ 10.3}   &{10.5 $\pm$ 0.2}   &{59.2 $\pm$ 8.7}   &{21.6 $\pm$ 2.4}   &\rc{48.0 $\pm$ 3.9}   &\rc{39.7 $\pm$ 3.9}    &{40.2}    \\
    \hline
    {Metasets~\cite{huang2021metasets}}       &\multirow{3}{*}{DG}                   &{47.4 $\pm$ 9.3}    &{14.0 $\pm$ 0.5} &{38.4 $\pm$ 12.2}  &{10.1 $\pm$ 0.8} &{21.8 $\pm$ 3.9}    &{19.6 $\pm$ 3.8} &{25.2}\\
    {PDG~\cite{wei2022learning}}              &                  &{31.9 $\pm$ 22.4}   &{24.4 $\pm$ 14.3}    &{43.8 $\pm$ 17.2}   &{14.0 $\pm$ 2.7}  &{30.6 $\pm$ 5.2}  &{30.5 $\pm$ 6.8}    &{29.2}\\
    {Ours}                                    &                  &\bb{66.2 $\pm$ 1.5}   &\bb{30.9 $\pm$ 2.2}   &\bb{81.7 $\pm$ 0.7}   &{30.1 $\pm$ 1.3}   &\bb{48.6 $\pm$ 2.3}  &\bb{39.8 $\pm$ 2.4}    &\bb{49.6}    \\
    \hline
    \hline
    \multicolumn{9}{c}{PointNet~\cite{qi2017pointnet}}\\
    \hline
    Supervised                    &\multirow{2}{*}{-}     &{70.1 $\pm$ 7.9}   &{40.9 $\pm$ 4.0}   &{86.4 $\pm$ 18.8}   &{40.9 $\pm$ 4.0}   &{86.4 $\pm$ 18.8}   &{70.1 $\pm$ 7.9}    &{65.8} \\
    w/o Adapt                     &                  &{51.8 $\pm$ 11.7}   &{21.9 $\pm$ 3.4}   &{67.5 $\pm$ 6.6}   &{24.5 $\pm$ 2.2}   &{33.8 $\pm$ 4.2}   &{32.9 $\pm$ 2.6}       &{38.7}   \\
    \hline
    VN~\cite{Deng_2021_ICCV}      &\multirow{2}{*}{RE}    &{57.3 $\pm$ 0.0}   &{22.0 $\pm$ 0.0}   &\rc{68.9 $\pm$ 0.0}   &{20.1 $\pm$ 0.0}   &\rc{34.5 $\pm$ 0.0}   &{26.3 $\pm$ 0.0}  &\rc{38.2} \\
    SVN~\cite{su2022svnet}      &    &{52.7 $\pm$ 0.2}   &{22.3 $\pm$ 0.3}   &{56.2 $\pm$ 0.2}   &{18.1 $\pm$ 0.0}   &{17.0 $\pm$ 0.3}   &{16.7 $\pm$ 0.2}  &{30.5} \\
    \hline
    
    % \hline
    PointDAN~\cite{qin2019pointdan} &\multirow{6}{*}{DA}   &{40.4 $\pm$ 10.4}   &{17.5 $\pm$ 2.5}   &{37.6 $\pm$ 7.4}   &{14.8 $\pm$ 3.5}   &{10.8 $\pm$ 0.5}   &{11.8 $\pm$ 0.8}    &{22.2}    \\
    DefRec~\cite{achituve2021self}  &                   &{34.8 $\pm$ 8.6}   &{15.0 $\pm$ 3.1}   &{35.3 $\pm$ 9.9}   &{14.3 $\pm$ 5.4}   &{13.9 $\pm$ 2.8}   &{11.8 $\pm$ 2.2}    &{20.9}     \\
    GAST~\cite{zou2021geometry}     &                   &{45.3 $\pm$ 4.1}   &\rc{24.6 $\pm$ 1.5}   &{53.6 $\pm$ 3.0}   &{20.3 $\pm$ 1.9}   &{21.4 $\pm$ 2.8}   &{24.4 $\pm$ 3.1}    &{31.6}     \\
    MLSP~\cite{liang2022point}      &                   &{44.2 $\pm$ 10.5}   &{21.8 $\pm$ 4.4}   &{46.6 $\pm$ 8.4}   &{18.3 $\pm$ 3.5}   &{20.2 $\pm$ 5.7}   &{19.0 $\pm$ 3.8}    &{28.4}     \\
    SDDA~\cite{cardace2023self}     &                   &\rc{58.4 $\pm$ 12.3}   &{19.0 $\pm$ 2.5}   &{55.0 $\pm$ 7.1}   &\rc{21.9 $\pm$ 3.7}   &{32.3 $\pm$ 6.9}   &{28.2 $\pm$ 4.5}    &{35.8}    \\
    PCFEA~\cite{wang2024progressive}     &                   &{50.2 $\pm$ 11.3}   &{10.4 $\pm$ 0.2}   &{44.8 $\pm$ 13.2}   &{10.4 $\pm$ 0.5}   &{17.1 $\pm$ 2.3}   &\rc{33.1 $\pm$ 3,2}    &{27.7}    \\
    \hline
    {Metasets~\cite{huang2021metasets}}&\multirow{3}{*}{DG}                   &{31.0 $\pm$ 7.3}    &{11.9 $\pm$ 0.9} &{32.5 $\pm$ 12.7}  &{10.0 $\pm$ 1.0} &{14.3 $\pm$ 4.5}    &{12.3 $\pm$ 2.4} &{18.7}\\
    {PDG~\cite{wei2022learning}}   &                  &{34.0 $\pm$ 11.5}   &{21.1 $\pm$ 3.0}    &{37.3 $\pm$ 17.8}   &{20.0 $\pm$ 4.2}  &{31.0 $\pm$ 8.9}  &{27.0 $\pm$ 5.3}    &{28.4}\\
    {Ours}                         &                    &\bb{63.3 $\pm$ 1.4}   &\bb{27.3 $\pm$ 1.3}   &\bb{81.4 $\pm$ 1.9}   &\bb{29.0 $\pm$ 1.2}   &\bb{42.7 $\pm$ 2.6}   &\bb{34.7 $\pm$ 1.5}       &\bb{46.4} \\
    \hline

    \end{tabular}

    }
    }
    \end{center}
    \vspace{-7mm}
\end{table*}

The entire framework is optimized in an alternative scheme between the intricate orientation mining and the orientation-aware contrastive training.
For the intricate orientation mining, the final objective is the same as Eq.~\ref{formula:iom}, where $L$ is the standard cross-entropy function over the originally aligned point clouds.
The intricate set is periodically updated after $T$ epochs of training to avoid overfitting to the current intricate rotations.
For orientation-aware contrastive training, the final loss function is
\begin{equation}
    L_{final} = L_{cls} + \lambda_{oc} L_{oc} + \lambda_{ms} L_{ms},
    \label{eq9}
\end{equation}
where $L_{cls}$ is the cross-entropy loss on the intricate point clouds. 

% \vspace{-3mm}
\section{Experiments} \label{sec:experiment}

\subsection{Experiment Setup}

\textbf{Datasets.}
We conduct experiments on the widely used 3d cross-domain benchmark PointDA~\cite{qin2019pointdan} and PointSegDA~\cite{achituve2021self} for shape classification and part segmentation, respectively. PointDA collects shapes of 10 categories from ModelNet (M)~\cite{wu20153d}, ShapeNet (S)~\cite{chang2015shapenet}, and ScanNet (S*)~\cite{dai2017scannet}, with all objects aligned in the up-right direction and down-sampled into 1024 points. We perform six cross-domain classification tasks on PointDA, including {M$\to$S}, {M$\to$S*}, {S$\to$M}, {S$\to$S*}, {S*$\to$M}, and {S*$\to$S}. Twelve cross-domain tasks are conducted on PointSegDA. This dataset consists of several human body shapes that are collected with different point distributions, poses, and scanned humans from four subdomains: Adobe, Faust, Mit, and Scape. The shapes share eight categories of human body parts (head, hand, feet, \etc) while down-sampling into 2048 points.
We strictly follow the same dataset splitting (80\% for training and 20\% for testing) for PointDA as~\cite{qin2019pointdan} and the same data split for PointSegDA as~\cite{achituve2021self} to implement our method under the orientation-aware 3D domain generalization setting.

\noindent\textbf{Evaluation and Metrics.}
To simulate the orientational shift, we randomly generate a series of SO(3) rotation matrices for each point cloud by uniformly sampling the rotation angles of the three-axises in the SO(3) space over $\left[-\pi, \pi\right]$, and rebuild the dataset with free-axis rotated augmentation for training. To comprehensively evaluate the performance and generalizability of each method under the orientation-aware 3D domain generalization setting, we sample 4 equidistant rotation angles per axis, \ie, $[\frac{\pi}{2}, \pi, \frac{3\pi}{2}, 2\pi]$, and construct 64 rotation augmented testing series. Each method is evaluated to obtain the average performance together with variance over the 64 tests. The statistics are presented by the mean of 5 times training for each method. We report the results of PointDA in the form of macro-average precision score (\textit{Avg.}) while measuring the mean Intersection-over-Union (\textit{mIoU}) for PointSegDA to evaluate the segmentation performance.

\noindent\textbf{Compared Methods.}
We compare our method with both 3DDG (Metasets~\cite{huang2021metasets}, PDG~\cite{wei2022learning}) and 3DDA (PointDAN~\cite{qin2019pointdan}, DefRec~\cite{achituve2021self}, GAST~\cite{zou2021geometry}, MLSP~\cite{liang2022point}, SDDA~\cite{cardace2023self}, and PCFEA~\cite{wang2024progressive}) methods on both cross-domain generalization classification and part segmentation tasks. Since they originally ignored the orientational shift, we obtained the experiment results by executing their official codes under the orientation-aware 3D domain generalization setting for fair comparison. To evaluate the robustness of our method towards orientational shift, we also compare with state-of-the-art rotation-equivalent and rotation-invariant point cloud analysis methods by applying them in the cross-domain scenario, including EOMP~\cite{luo2022equivariant}, VN~\cite{Deng_2021_ICCV}, SVN~\cite{su2022svnet}, SPRIN~\cite{you2021prin}, RIPCA~\cite{li2021closer}, RIConv++~\cite{zhang2022riconv}, PaRI~\cite{chen2022devil}, and LocoTrans~\cite{chen2024local}.

\noindent\textbf{Implementation Details.}
We implement our framework in Fig.~\ref{fig:pipeline} with the widely used rotation-invariant model PointNet~\cite{qi2017pointnet} and non-invariant model DGCNN~\cite{wang2019dynamic} as our backbones, training with 200 epochs and batch size 32 on a single RTX3090 GPU. We employ Adam~\cite{kingma2014adam} as the optimizer. The update period $T$ of the intricate orientation set is set as 20. The learning rate is initialized to $10^{-3}$ with a decay weight of $10^{-4}$ and scheduled by a degradation function $(1 + \gamma (ep/ep_{max}))^{-\beta}$ during the training phase, where $ep_{max}$ is the maximum number of training epochs, and $\gamma, \beta$ are set as 10 and 0.75, respectively. The temperature factor $\tau_s$, $\tau_t$ in Eq.~\ref{eq5} are empirically set to 0.5 and $\tau'$ in Eq.~\ref{eq6} and Eq.~\ref{eq7} is set to 0.07. 
The number of repeat times $AT$ is set as 10, while the number of the intricate variants $V$ in Eq.~\ref{eq6} are set as 5 due to limited memory. $\lambda_{oc}$ and $\lambda_{ms}$ in Eq.~\ref{eq9} are both set as 0.01 by observing the performance changes on M$\to$S (Please refer to the hyper-parameter sensitivity analysis in the supplementary material.). For part segmentation, we simply equip the backbone networks in our framework with an extra decoder for dense prediction and extend the $L_{cls}$ in Eq.~\ref{eq9} into a pixel-wise cross-entropy loss. All our experiments share the same set of hyperparameters. 
For the compared SDDG and SDDA methods, we trained their models with the vanilla architectures of PointNet/DGCNN. For rotation-equivalent methods and rotation-invariant methods (\eg, EOMP~\cite{luo2022equivariant}, SPRIN~\cite{you2021prin}, and RIPCA~\cite{li2021closer}), we strictly follow their official designs on network architectures and augmentation strategies without modifications. This adherence is crucial as the unique modules proposed in these methods are integral to maintaining the rotational invariance of point features.
The augmentation strategies we adopted are the same as~\cite{qin2019pointdan}, including random points down-sampling, random rotation, and point jittering, which are also used in the compared methods except for Metasets and PDG, as we empirically found that maintaining their original augmentation setting yields the best performance for these two methods. 

\begin{table*}[h] % table for segmentation task on PointSegDA
    \caption{Comparison of the mIoU under the orientation-aware 3D domain generalization part segmentation setting using DGCNN. The value after $\pm$ denotes the standard deviation on the 64 evaluated series. The top two records are marked in \bc{red} and \rc{blue}, respectively.}
    \label{tab:pointda10_mIoU} 
    \vspace{-4mm}
    \small
    \begin{center}
    \setlength{\tabcolsep}{0.03cm}{
    \resizebox{1\textwidth}{!}{
    \begin{tabular}{l|c|l|l|l|l|l|l|l|l|l|l|l|l|l}
    \hline

    \makecell[c]{Methods}
    &Type &\makecell[c]{Adobe to \\ Faust} & \makecell[c]{Adobe to \\ Mit} & \makecell[c]{Adobe to \\ Scape}  &\makecell[c]{Faust to \\ Adobe} & \makecell[c]{Faust to \\ Mit} & \makecell[c]{Faust to \\ Scape}  &\makecell[c]{Mit to \\ Adobe} & \makecell[c]{Mit to \\ Faust} & \makecell[c]{Mit to \\ Scape}  & \makecell[c]{Scape to \\ Adobe} & \makecell[c]{Scape to \\ Faust} & \makecell[c]{Scape to \\ Mit} & {AVG} \\
    \hline
    Supervised               &\multirow{2}{*}{-}     &{ 81.2 $\pm$ 14.1}   &{ 76.3 $\pm$ 16.1}   &{ 79.6 $\pm$ 20.1}   &{ 80.8 $\pm$ 14.2}   &{ 76.3 $\pm$ 16.1}   &{ 79.6 $\pm$ 20.1}   &{ 80.8 $\pm$ 14.2}   &{ 81.2 $\pm$ 14.1} &{ 79.6 $\pm$ 20.1}   &{ 80.8 $\pm$ 14.2}   &{ 81.2 $\pm$ 14.1}   &{ 76.3 $\pm$ 16.1}  &{ 79.5} \\
    w/o Adapt                     &                  &{ 39.8 $\pm$ 9.5}   &{ 22.9 $\pm$ 5.0}   &{ 31.6 $\pm$ 7.3}   &{ 62.9 $\pm$ 19.8}   &{ 47.9 $\pm$ 14.3}   &{ 58.8 $\pm$ 17.3}  &{ 37.0 $\pm$ 13.3}   &{ 34.9 $\pm$ 9.4}  &{ 63.2 $\pm$ 14.1}   &{ 52.4 $\pm$ 14.7}   &{ 56.2 $\pm$ 15.5}   &{ 51.4 $\pm$ 15.7}  &{ 46.6}   \\
    \hline
    VN~\cite{Deng_2021_ICCV}      &\multirow{3}{*}{RE}    &{ 43.4 $\pm$ 0.0}   &{ 21.2 $\pm$ 0.0}   &{ 28.5 $\pm$ 0.0}   &{ 64.7 $\pm$ 0.0}   &{ 18.6 $\pm$ 0.0}   &{ 30.2 $\pm$ 0.0}  &{ 31.7 $\pm$ 0.0}   &{ 24.4 $\pm$ 0.0}  &{ 43.4 $\pm$ 0.0}   &{ 25.0 $\pm$ 0.0}   &{ 30.1 $\pm$ 0.0}   &{ 38.9 $\pm$ 0.0}  &{ 33.3} \\
    SVN~\cite{su2022svnet}      &    &{ 33.6 $\pm$ 0.0}   &{ 17.3 $\pm$ 0.0}   &{ 21.6 $\pm$ 0.0}   &{ 16.1 $\pm$ 0.0}   &{ 10.4 $\pm$ 0.0}   &{ 20.2 $\pm$ 0.0}  &{ 20.6 $\pm$ 0.0}   &{ 17.4 $\pm$ 0.0}   &{ 17.2 $\pm$ 0.0}   &{ 18.1 $\pm$ 0.0}   &{ 17.0 $\pm$ 0.3}   &{ 16.7 $\pm$ 0.2} &{ 30.5} \\
    EOMP~\cite{luo2022equivariant}            &      & \bc{ 48.5 $\pm$ 0.1}    &{ 14.6 $\pm$ 0.0} &{ 32.4 $\pm$ 0.0}  &{ 64.2 $\pm$ 0.0} &{ 24.9 $\pm$ 0.1}    &{ 63.6 $\pm$ 0.1} &{ 48.6 $\pm$ 0.2}    &{ 38.4 $\pm$ 0.0} &{ 32.6 $\pm$ 0.1}  &{ 34.1 $\pm$ 0.1} &{ 11.8 $\pm$ 0.0}    &{ 23.4 $\pm$ 0.1}&{ 36.4}\\ \cline{2-2}
    SPRIN~\cite{you2021prin}    &\multirow{3}{*}{RI}    &{ 42.6 $\pm$ 0.2}   &\rc{ 23.5 $\pm$ 0.4}   &{ 31.9 $\pm$ 0.8}   & \bc{ 78.9 $\pm$ 1.3}   &{ 33.8 $\pm$ 0.6}   &\rc{ 67.1 $\pm$ 1.3}  &{ 42.9 $\pm$ 0.9}   &\rc{ 55.7 $\pm$ 0.8}   &{ 63.8 $\pm$ 0.8}   & \bc{ 70.8 $\pm$ 0.8}   & \bc{ 77.8 $\pm$ 0.5}   &\rc{ 60.6 $\pm$ 1.1}  &\rc{ 54.1} \\
    RIPCA~\cite{li2021closer}              &            &{ 45.5 $\pm$ 1.0} &{ 20.7 $\pm$ 2.6}   &\rc{ 34.6 $\pm$ 2.0} &{ 65.5 $\pm$ 2.1}    &\rc{ 47.6 $\pm$ 1.9} &{ 65.9 $\pm$ 1.5}  &\rc{ 50.9 $\pm$ 3.9} &{ 50.4 $\pm$ 4.6}   &\rc{ 72.9 $\pm$ 1.3} &{ 54.9 $\pm$ 1.7}    &{ 56.3 $\pm$ 2.5} &{ 53.7 $\pm$ 6.0} &{ 51.6}\\
    LocoTrans~\cite{chen2024local}              &            &{ 43.7 $\pm$ 0.0}  &{ 20.6 $\pm$ 0.0}  &{ 29.5 $\pm$ 0.0}  &{ 67.3 $\pm$ 0.0}  &{ 33.1 $\pm$ 0.0}  &{ 47.6 $\pm$ 0.0}  &{ 22.7 $\pm$ 0.0}  &{ 39.8 $\pm$ 0.0}  &{ 67.1 $\pm$ 0.0}  &{ 51.7 $\pm$ 0.0}  &{ 65.2 $\pm$ 0.0}  &{ 62.0 $\pm$ 0.0} &{ 45.9}\\
    \cline{2-2}
    DefRec~\cite{achituve2021self}  &\multirow{3}{*}{DA}                   &{ 40.0 $\pm$ 9.1}   &{ 22.5 $\pm$ 6.8}   &{ 31.4 $\pm$ 8.4}   &{ 38.6 $\pm$ 11.0}   &{ 34.8 $\pm$ 14.8}   &{ 46.8 $\pm$ 16.1}  &{ 34.0 $\pm$ 11.1}   &{ 46.2 $\pm$ 11.3}   &{ 53.3 $\pm$ 16.8}   &{ 44.0 $\pm$ 11.6}   &{ 49.2 $\pm$ 12.4}   &{ 46.1 $\pm$ 13.2}  &{ 40.6}     \\
    MLSP~\cite{liang2022point}      &                   &{ 38.7 $\pm$ 6.8}   &{ 20.3 $\pm$ 4.1}   &{ 29.9 $\pm$ 5.2}   &{ 45.1 $\pm$ 13.9}   &{ 43.6 $\pm$ 12.5}   &{ 57.7 $\pm$ 17.0}    &{ 24.8 $\pm$ 8.8}   &{ 38.9 $\pm$ 12.2}   &{ 55.1 $\pm$ 17.3}   &{ 36.2 $\pm$ 12.4}   &{ 54.5 $\pm$ 15.2}   &{ 45.2 $\pm$ 13.8} &{ 40.8}     \\
    SDDA~\cite{cardace2023self}    &                   &{ 33.5 $\pm$ 11.0}   &{ 18.3 $\pm$ 5.5}   &{ 27.4 $\pm$ 9.3}   &{ 32.8 $\pm$ 16.4}   &{ 38.2 $\pm$ 11.8}   &{ 41.9 $\pm$ 16.7}    &{ 29.1 $\pm$ 11.1}   &{ 39.1 $\pm$ 12.4}   &{ 43.5 $\pm$ 16.2}   &{ 36.4 $\pm$ 14.5}   &{ 47.5 $\pm$ 15.6}   &{ 39.9 $\pm$ 12.8} &{ 35.6}     \\
    \cline{2-2}
    {Ours}                         &DG                    &\rb{ 47.7 $\pm$ 0.9}   &\bb{ 27.5 $\pm$ 2.5}    &\bb{ 39.7 $\pm$ 2.4}   &\rb{ 73.7 $\pm$ 2.7}   &\bb{ 60.8 $\pm$ 3.1}   &\bb{ 74.5 $\pm$ 1.1}   &\bb{ 56.5 $\pm$ 3.6}   &\bb{ 56.4 $\pm$ 3.1}   &\bb{ 78.3 $\pm$ 1.3}   &\rb{ 65.1 $\pm$ 2.7}   &\rb{ 74.8 $\pm$ 1.0}   &\bb{ 72.6 $\pm$ 1.9}    &\bb{ 60.6} \\
    \hline

    \end{tabular}

    }
    }
    \end{center}
    \vspace{-7mm}
\end{table*}

\vspace{-2mm}
\subsection{Experimental Results}

In this section, we report the compared results on classification (Table~\ref{tab:pointda10_avg}) and part segmentation tasks (Table~\ref{tab:pointda10_mIoU}), respectively. The values after $\pm$ denote the standard deviation over the 64 rotation tests, indicating the consistency of the evaluation results towards orientational shift. The smaller the values are, the higher the stability of recognition concerning various orientations.

\noindent\textbf{Orientation-aware 3D Domain Generalization for Classification.}
By averaging the performance over the six domain generalization tasks, our method achieves 8.3\% and 7.7\% improvements on the \textit{Avg.} compared to the DGCNN and the PointNet baselines, respectively, and also outperforms all other methods compared.
As shown in Table~\ref{tab:pointda10_avg}, our framework achieves the best classification results in five out of the six tasks using DGCNN as the backbone, which exceeds both the current state-of-the-art 3D domain generation and rotation-invariant methods by a large margin. When adopting PointNet as the backbone, our performance even exceeds all the other methods over the six cross-domain tasks, demonstrating a more balanced discriminability. Additionally, the relatively small values of the standard deviations across different rotations also validate our model's consistency \wrt rotations.

\begin{table}[t]
    \caption{Comparison among variants of our method on M$\to$S with different backbones.}
    \vspace{-2mm}
    \label{tab:ablation} 
    \small
    \centering
    
    \setlength{\tabcolsep}{0.2cm}{
    \begin{tabular}{c|ccc|cc|cc}
    \hline

    \multirow{2}{*}{Variants} &\multirow{2}{*}{IOM} &\multirow{2}{*}{OC}  &\multirow{2}{*}{MS} &\multicolumn{2}{c|}{DGCNN~\cite{wang2019dynamic}} &\multicolumn{2}{c}{PointNet~\cite{qi2017pointnet}} \\
    \cline{5-8}
    &&&                    &{Avg.}  &Cst. &{Avg.} &Cst.\\
    \hline
    baseline &{    } &{    } &{    }                    &{54.7}  &{1.279} &{51.8} &{1.404}\\
    V1 &{$\checkmark$} &{    } &{    }                  &{56.0}  &{0.446} &{55.9} &{0.532}\\
    V2 &{$\checkmark$} &{$\checkmark$} &{    }          &{60.2}  &\tb{0.392} &{57.4} &\tb{0.447}\\ 
    V3 &{$\checkmark$} &{    } &{$\checkmark$}           &{55.3} &{0.552} &{54.3} &{0.600}\\ 
    V4 &{    } &{$\checkmark$} &{$\checkmark$}          &{48.8}  &{0.503} &{52.1} &{0.628}\\ 
    Ours &{$\checkmark$} &{$\checkmark$} &{$\checkmark$}     &\tb{66.2}  &{0.425} &\tb{63.3} &{0.456}\\
    \hline

    \end{tabular}
    }
    \vspace{-5.5mm}
\end{table}

\noindent\textbf{Orientation-aware 3D Domain Generalization for Part Segmentation.}
For part segmentation, we compare with several outstanding competitors in Table~\ref{tab:pointda10_avg}, and the results are presented in Table~\ref{tab:pointda10_mIoU}. Our method achieves a 6.5\% increase in the average \textit{mIoU} over the twelve part segmentation tasks. More concretely, our method achieves the best segmentation quality in eight out of the twelve scenes and the second-best results in the rest four scenes, demonstrating its flexibility for a broader range of cross-domain point cloud analysis applications beyond classification.

\noindent\textbf{Comparison with SDDG and SDDA Methods. }
Compared to other 3DDG works, our method demonstrates significant superiority, with 20.4\% and 24.4\% improvement in \textit{Avg.} over PDG and Metasets, respectively. Metasets are more prone to overfitting under the orientational shift as shown in Table~\ref{tab:pointda10_avg}, which can be attributed to the heavy augmentations during meta-training.
We also notice that the performance of PDG is much more unstable and even inferior to the baseline.
The possible reason is that the part-based representation adopted by PDG concentrates on the local geometric features, which are more delicate under the shift of rotation.
Compared with 3DDA methods, our method still shows great power even though they accessed the target data during the training phase. Given MLSP as a reference, our method achieves prominent performance gains, with an improvement of 6.4\% on \textit{Avg.} and 19.8\% on \textit{mIoU}, respectively. This phenomenon is subject to the design of most state-of-the-art 3DDA methods. They focus on learning the geometric information of shape via self-supervised tasks, which are highly sensitive to rotation priors. Generally, all the 3DDA methods have large variances on different rotated test sequences, which indicates that they are less stable under the perturbation of arbitrary orientations. For GAST, which highly relies on accurate supervision of point normals, the performance deteriorates as the error in estimating normals is amplified by unpredictable rotations.

\noindent\textbf{Comparison with RE and RI Methods.} The RE and RI methods are powerful in solving in-domain rotation ambiguity, however, they lack further consideration of the cross-domain scene. This may result in a drastic performance drop when applying the model to other scenes. As shown in Tables~\ref{tab:pointda10_avg} and~\ref{tab:pointda10_mIoU}, the performance of our method in both cross-domain classification and part segmentation still exceeds existing RE and RI methods. Our method achieves 3.0\% improvement in \textit{Avg.} over the best RI classification method RIPCA, and 6.5\% improvement in \textit{mIoU} compared to the best method SPRIN for part segmentation. The standard deviations of our method are also comparable to these methods, indicating considerable rotational robustness.

\vspace{-1mm}
\subsection{Ablation Study and Analysis}

In this section, we delve into the effectiveness of each proposed component of our method, including the intricate orientation mining (IOM), the orientation consistent loss (OC), and the margin separation loss (MS). The ablations are conducted on M$\to$S using both DGCNN and PointNet as our backbones. We also visualize the effects of different values of $\lambda_{oc}$ and $\lambda_{ms}$ in Eq.~\ref{eq9}. Additionally, we investigate the efficacy of intricate orientation mining during the optimization process and provide an evaluation of the time complexity of our method.

\noindent\textbf{Effectiveness of Each Component.} We conduct ablation studies by designing four variants on different backbones, including directly disabling part of the components for V1 to V3 and replacing the IOM with random rotation augmentation in V4. Apart from the \textit{Acc.} and \textit{Avg.}, an extra metric (\textit{Cst.}) is reported to reflect the consistency of the model. For each metric, we report the expectation value of the whole testing set. Specifically, we augment each point cloud 64 times and calculate the \textit{Cst.} by measuring the mean KL-divergence over their output probabilities. The smaller the \textit{Cst.} is, the more consistent the model will be. As illustrated in Table~\ref{tab:ablation}, directly training with intricate samples from intricate orientation mining can significantly enhance the consistency of the model (baseline vs. V1). Incorporating inner-sample contrastive learning on samples with intricate orientation augmentation (V1 vs. V2) could further improve the consistency and promote features's expressiveness that are beneficial for classification. MS collaborates with the representative features learned by OC and further enhances discriminability by explicitly increasing the margin between each category (V2 vs. Ours), however, solely applying MS leads to inconsistent predictions of multiple point cloud's rotated variants, which increases the difficulty of learning a robust classifier (V1 vs. V3). Finally, the overfitting problem of random rotation is evidenced by V4, which intensifies the biased learning problem by applying OC and MS directly on the random rotated point clouds (V4 vs. Ours), demonstrating the effectiveness of IOM in tackling rotational shift under the 3D domain generalization setting.
\begin{figure}[t]
    % \flushleft
    \centering
    \subfloat[intricate orientations mining only]{%[b]{0.45\textwidth}
        \label{fig:feat_iom}
        % \centering
        \includegraphics[width=0.24\textwidth]{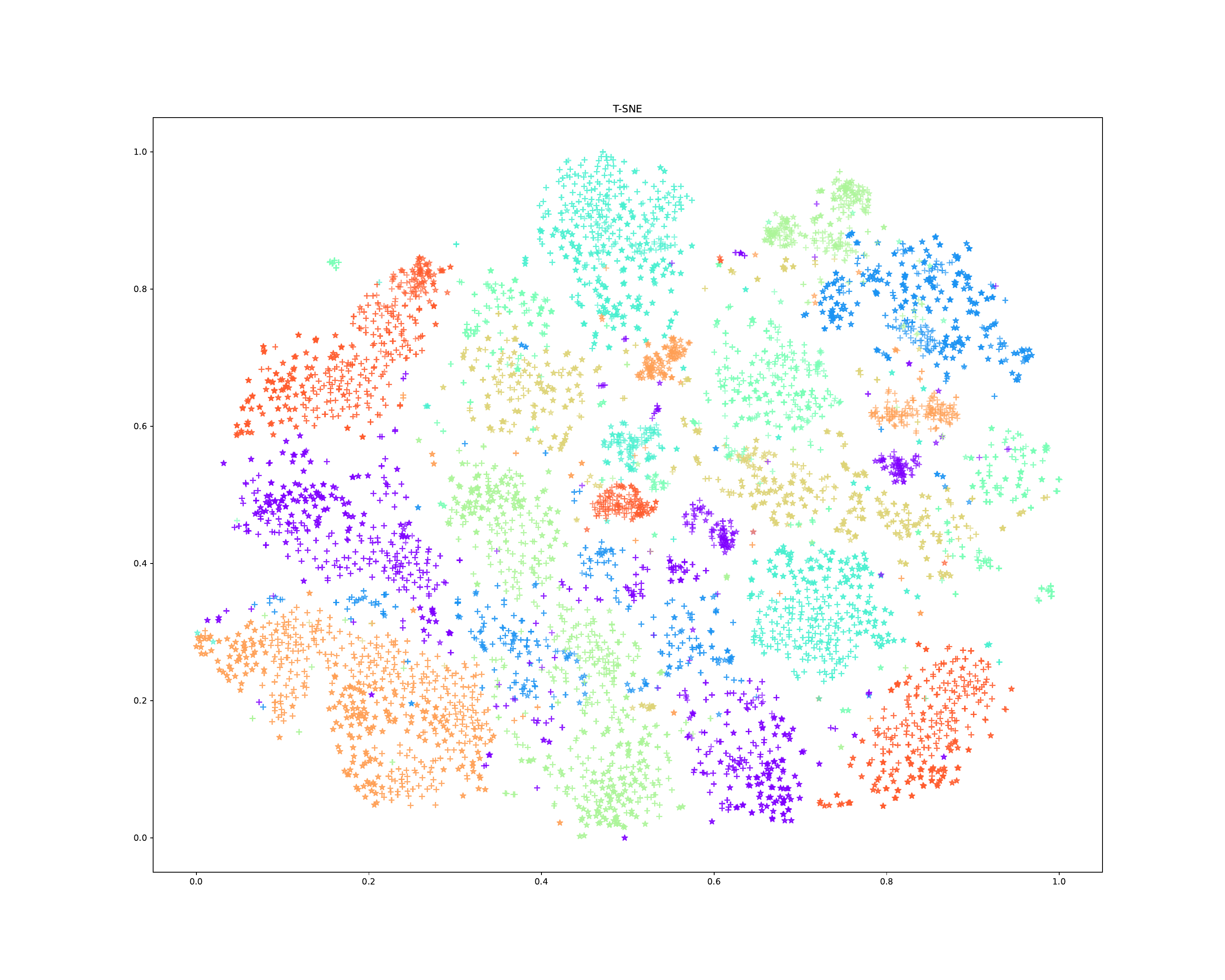}
    }
    % \hspace{.4in}
    \subfloat[orientation-aware contrastive training]{%[b]{0.45\textwidth}
        \label{fig:feat_all}
        % \centering
        \includegraphics[width=0.24\textwidth]{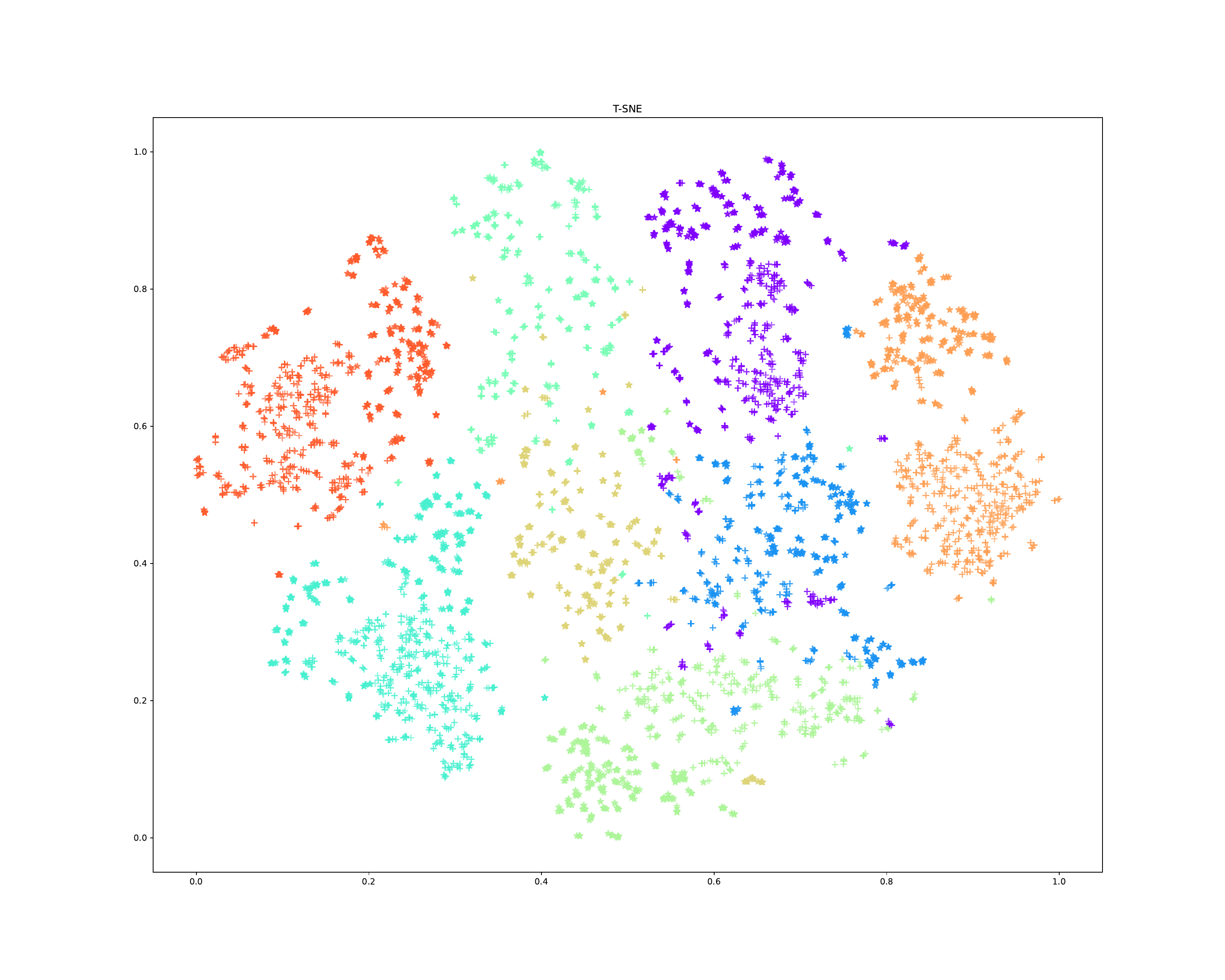}
    }
    \vspace{-2mm}
    \caption{t-SNE visualizations of the feature space (a) trained using only intricate orientation mining, and (b) further refined by orientation-aware contrastive training. Point features from ModelNet (the source domain) and ShapeNet (the target domain) are denoted by star (`$\star$') and plus (`$+$'), respectively.}
    \label{fig:feature}
    \vspace{-6mm}
\end{figure}

\noindent\textbf{Visualization of Point Cloud Features.} 
For a more comprehensive understanding of the effects of the proposed orientation-aware contrastive training on improving the generalizability of learned point cloud features, in Fig.~\ref{fig:feature}, we employ t-SNE~\cite{van2008visualizing} to visualize the feature spaces generated by our proposed method before and after applying the orientation-aware contrastive training (V1 vs. Ours) in the M$\to$S experiment. For each subdomain, we select the first 50 samples from each category and augment them by uniformly sampling 10 out of the 64 testing rotation series.

The features extracted using V1 are depicted in Fig.~\ref{fig:feature}(a), where we can observe that despite training with the intricately augmented version of each point cloud, the resulting feature space remains non-separable, with features of the same category fragmenting into multiple small clusters, such as the orange and purple points. However, notable spatial proximity can be observed \wrt the source features (`$\star$') and the target features (`$+$') in the feature space. This phenomenon corroborates our findings in Section~\ref{sec:analysis}: while augmenting with intricate orientations mitigates the domain gap induced by arbitrary rotations, there is still significant room for enhancing feature discriminability and generalizability. 
As revealed in Fig.~\ref{fig:feature}(b), upon integrating our orientational-aware contrastive training, the resulting feature space exhibits enhanced discriminability. The consistency between a given point cloud and its rotated variants is also markedly improved. Compared to V1, the feature clusters of the same category from both source and target domains become closer, with more clear boundaries between adjacent categories. The feature spaces of both domains are now more separable, benefiting the learning of a more generalizable classifier in cross-domain scenarios.

\vspace{-2mm}
\section{Conclusion} \label{sec:conclusion}
We present a comprehensive exploration of the orientational shift in 
the cross-domain 3D classification task. We observe that the intricate orientations have advantageous impact on improving generalizability. A novel intricate orientation learning framework is proposed to tackle the orientation-aware 3D domain generalization problem, which empowers the model with enhanced rotational consistency and discriminability. Extensive experiments demonstrate the superiority of our method.

{\small
\bibliographystyle{ieee_fullname}
\bibliography{egbib}

\begin{thebibliography}{10}\itemsep=-1pt

\bibitem{achituve2021self}
Idan Achituve, Haggai Maron, and Gal Chechik.
\newblock Self-supervised learning for domain adaptation on point clouds.
\newblock In {\em WACV}, pages 123--133, 2021.

\bibitem{borgwardt2006integrating}
Karsten~M Borgwardt, Arthur Gretton, Malte~J Rasch, Hans-Peter Kriegel,
  Bernhard Sch{\"o}lkopf, and Alex~J Smola.
\newblock Integrating structured biological data by kernel maximum mean
  discrepancy.
\newblock {\em Bioinformatics}, 22(14):e49--e57, 2006.

\bibitem{brostow2008segmentation}
Gabriel~J Brostow, Jamie Shotton, Julien Fauqueur, and Roberto Cipolla.
\newblock Segmentation and recognition using structure from motion point
  clouds.
\newblock In {\em ECCV}, pages 44--57. Springer, 2008.

\bibitem{cardace2023self}
Adriano Cardace, Riccardo Spezialetti, Pierluigi~Zama Ramirez, Samuele Salti,
  and Luigi Di~Stefano.
\newblock Self-distillation for unsupervised 3d domain adaptation.
\newblock In {\em WACV}, pages 4166--4177, 2023.

\bibitem{chang2015shapenet}
Angel~X Chang, Thomas Funkhouser, Leonidas Guibas, Pat Hanrahan, Qixing Huang,
  Zimo Li, Silvio Savarese, Manolis Savva, Shuran Song, Hao Su, et~al.
\newblock Shapenet: An information-rich 3d model repository.
\newblock {\em arXiv preprint arXiv:1512.03012}, 2015.

\bibitem{chen2019clusternet}
Chao Chen, Guanbin Li, Ruijia Xu, Tianshui Chen, Meng Wang, and Liang Lin.
\newblock Clusternet: Deep hierarchical cluster network with rigorously
  rotation-invariant representation for point cloud analysis.
\newblock In {\em CVPR}, pages 4994--5002, 2019.

\bibitem{chen2021equivariant}
Haiwei Chen, Shichen Liu, Weikai Chen, Hao Li, and Randall Hill.
\newblock Equivariant point network for 3d point cloud analysis.
\newblock In {\em CVPR}, pages 14514--14523, 2021.

\bibitem{chen2022devil}
Ronghan Chen and Yang Cong.
\newblock The devil is in the pose: Ambiguity-free 3d rotation-invariant
  learning via pose-aware convolution.
\newblock In {\em CVPR}, pages 7472--7481, 2022.

\bibitem{chen2024local}
Yiyang Chen, Lunhao Duan, Shanshan Zhao, Changxing Ding, and Dacheng Tao.
\newblock Local-consistent transformation learning for rotation-invariant point
  cloud analysis.
\newblock In {\em CVPR}, pages 5418--5427, 2024.

\bibitem{dai2017scannet}
Angela Dai, Angel~X Chang, Manolis Savva, Maciej Halber, Thomas Funkhouser, and
  Matthias Nie{\ss}ner.
\newblock Scannet: Richly-annotated 3d reconstructions of indoor scenes.
\newblock In {\em CVPR}, pages 5828--5839, 2017.

\bibitem{Deng_2021_ICCV}
Congyue Deng, Or Litany, Yueqi Duan, Adrien Poulenard, Andrea Tagliasacchi, and
  Leonidas~J. Guibas.
\newblock Vector neurons: A general framework for so(3)-equivariant networks.
\newblock In {\em ICCV}, pages 12200--12209, 2021.

\bibitem{dollar2009integral}
Piotr Doll{\'a}r, Zhuowen Tu, Pietro Perona, and Serge~J Belongie.
\newblock Integral channel features.
\newblock In {\em BMVC}, volume~2, page~5, 2009.

\bibitem{fan2022self}
Hehe Fan, Xiaojun Chang, Wanyue Zhang, Yi Cheng, Ying Sun, and Mohan
  Kankanhalli.
\newblock Self-supervised global-local structure modeling for point cloud
  domain adaptation with reliable voted pseudo labels.
\newblock In {\em CVPR}, pages 6377--6386, 2022.

\bibitem{felzenszwalb2009object}
Pedro~F Felzenszwalb, Ross~B Girshick, David McAllester, and Deva Ramanan.
\newblock Object detection with discriminatively trained part-based models.
\newblock {\em IEEE TPAMI}, 32(9):1627--1645, 2009.

\bibitem{gretton2012kernel}
Arthur Gretton, Karsten~M Borgwardt, Malte~J Rasch, Bernhard Sch{\"o}lkopf, and
  Alexander Smola.
\newblock A kernel two-sample test.
\newblock {\em JMLR}, 13(1):723--773, 2012.

\bibitem{hinton2015distilling}
Geoffrey Hinton, Oriol Vinyals, and Jeff Dean.
\newblock Distilling the knowledge in a neural network.
\newblock {\em arXiv preprint arXiv:1503.02531}, 2015.

\bibitem{huang2021metasets}
Chao Huang, Zhangjie Cao, Yunbo Wang, Jianmin Wang, and Mingsheng Long.
\newblock Metasets: Meta-learning on point sets for generalizable
  representations.
\newblock In {\em CVPR}, pages 8863--8872, 2021.

\bibitem{kim2020rotation}
Seohyun Kim, Jaeyoo Park, and Bohyung Han.
\newblock Rotation-invariant local-to-global representation learning for 3d
  point cloud.
\newblock {\em NeurIPS}, 33, 2020.

\bibitem{kingma2014adam}
Diederik~P Kingma and Jimmy Ba.
\newblock Adam: A method for stochastic optimization.
\newblock {\em arXiv preprint arXiv:1412.6980}, 2014.

\bibitem{leibe2007dynamic}
Bastian Leibe, Nico Cornelis, Kurt Cornelis, and Luc Van~Gool.
\newblock Dynamic 3d scene analysis from a moving vehicle.
\newblock In {\em CVPR}, pages 1--8, 2007.

\bibitem{li2021closer}
Feiran Li, Kent Fujiwara, Fumio Okura, and Yasuyuki Matsushita.
\newblock A closer look at rotation-invariant deep point cloud analysis.
\newblock In {\em ICCV}, pages 16218--16227, 2021.

\bibitem{li2021rotation}
Xianzhi Li, Ruihui Li, Guangyong Chen, Chi-Wing Fu, Daniel Cohen-Or, and
  Pheng-Ann Heng.
\newblock A rotation-invariant framework for deep point cloud analysis.
\newblock {\em IEEE TVCG}, 28(12):4503--4514, 2021.

\bibitem{liang2022point}
Hanxue Liang, Hehe Fan, Zhiwen Fan, Yi Wang, Tianlong Chen, Yu Cheng, and
  Zhangyang Wang.
\newblock Point cloud domain adaptation via masked local 3d structure
  prediction.
\newblock In {\em ECCV}, pages 156--172. Springer, 2022.

\bibitem{lin2017focal}
Tsung-Yi Lin, Priya Goyal, Ross Girshick, Kaiming He, and Piotr Doll{\'a}r.
\newblock Focal loss for dense object detection.
\newblock In {\em ICCV}, pages 2980--2988, 2017.

\bibitem{luo2022equivariant}
Shitong Luo, Jiahan Li, Jiaqi Guan, Yufeng Su, Chaoran Cheng, Jian Peng, and
  Jianzhu Ma.
\newblock Equivariant point cloud analysis via learning orientations for
  message passing.
\newblock In {\em CVPR}, pages 18932--18941, 2022.

\bibitem{madrytowards}
Aleksander Madry, Aleksandar Makelov, Ludwig Schmidt, Dimitris Tsipras, and
  Adrian Vladu.
\newblock Towards deep learning models resistant to adversarial attacks.
\newblock In {\em CVPR}, 2017.

\bibitem{park2023pcadapter}
Joonhyung Park, Hyunjin Seo, and Yang Eunho.
\newblock Pc-adapter: Topology-aware adapter for efficient domain adaption on
  point clouds with rectified pseudo-label.
\newblock In {\em ICCV}, pages 11530--11540, 2023.

\bibitem{poulenard2021functional}
Adrien Poulenard and Leonidas~J Guibas.
\newblock A functional approach to rotation equivariant non-linearities for
  tensor field networks.
\newblock In {\em CVPR}, pages 13174--13183, 2021.

\bibitem{qi2017pointnet}
Charles~R Qi, Hao Su, Kaichun Mo, and Leonidas~J Guibas.
\newblock Pointnet: Deep learning on point sets for 3d classification and
  segmentation.
\newblock In {\em CVPR}, pages 652--660, 2017.

\bibitem{qin2019pointdan}
Can Qin, Haoxuan You, Lichen Wang, C-C~Jay Kuo, and Yun Fu.
\newblock Pointdan: A multi-scale 3d domain adaption network for point cloud
  representation.
\newblock {\em NeurIPS}, 32, 2019.

\bibitem{schroff2015facenet}
Florian Schroff, Dmitry Kalenichenko, and James Philbin.
\newblock Facenet: A unified embedding for face recognition and clustering.
\newblock In {\em CVPR}, pages 815--823, 2015.

\bibitem{su2022svnet}
Zhuo Su, Max Welling, Matti Pietik{\"a}inen, and Li Liu.
\newblock Svnet: Where so (3) equivariance meets binarization on point cloud
  representation.
\newblock In {\em 3DV}, pages 547--556. IEEE, 2022.

\bibitem{tarvainen2017mean}
Antti Tarvainen and Harri Valpola.
\newblock Mean teachers are better role models: Weight-averaged consistency
  targets improve semi-supervised deep learning results.
\newblock {\em NeurIPS}, 30, 2017.

\bibitem{van2008visualizing}
Laurens Van~der Maaten and Geoffrey Hinton.
\newblock Visualizing data using t-sne.
\newblock 9(11), 2008.

\bibitem{wang2021instance}
Weilun Wang, Wengang Zhou, Jianmin Bao, Dong Chen, and Houqiang Li.
\newblock Instance-wise hard negative example generation for contrastive
  learning in unpaired image-to-image translation.
\newblock In {\em ICCV}, pages 14020--14029, 2021.

\bibitem{wang2019dynamic}
Yue Wang, Yongbin Sun, Ziwei Liu, Sanjay~E Sarma, Michael~M Bronstein, and
  Justin~M Solomon.
\newblock Dynamic graph cnn for learning on point clouds.
\newblock {\em ACM TOG}, 38(5):1--12, 2019.

\bibitem{wang2024progressive}
Zicheng Wang, Zhen Zhao, Yiming Wu, Luping Zhou, and Dong Xu.
\newblock Progressive classifier and feature extractor adaptation for
  unsupervised domain adaptation on point clouds.
\newblock In {\em ECCV}, pages 37--55. Springer, 2024.

\bibitem{wei2022learning}
Xin Wei, Xiang Gu, and Jian Sun.
\newblock Learning generalizable part-based feature representation for 3d point
  clouds.
\newblock {\em NeurIPS}, 35, 2022.

\bibitem{wu20153d}
Zhirong Wu, Shuran Song, Aditya Khosla, Fisher Yu, Linguang Zhang, Xiaoou Tang,
  and Jianxiong Xiao.
\newblock 3d shapenets: A deep representation for volumetric shapes.
\newblock In {\em CVPR}, pages 1912--1920, 2015.

\bibitem{you2021prin}
Yang You, Yujing Lou, Ruoxi Shi, Qi Liu, Yu-Wing Tai, Lizhuang Ma, Weiming
  Wang, and Cewu Lu.
\newblock Prin/sprin: On extracting point-wise rotation invariant features.
\newblock {\em IEEE TPAMI}, 44(12):9489--9502, 2021.

\bibitem{yu2019unsupervised}
Hong-Xing Yu, Wei-Shi Zheng, Ancong Wu, Xiaowei Guo, Shaogang Gong, and
  Jian-Huang Lai.
\newblock Unsupervised person re-identification by soft multilabel learning.
\newblock In {\em CVPR}, pages 2148--2157, 2019.

\bibitem{zhang2022self}
Zaiwei Zhang, Min Bai, and Erran~Li Li.
\newblock Self-supervised pretraining for large-scale point clouds.
\newblock {\em NeurIPS}, 35, 2022.

\bibitem{zhang2020global}
Zhiyuan Zhang, Binh-Son Hua, Wei Chen, Yibin Tian, and Sai-Kit Yeung.
\newblock Global context aware convolutions for 3d point cloud understanding.
\newblock In {\em 3DV}, pages 210--219. IEEE, 2020.

\bibitem{zhang2022riconv}
Zhiyuan Zhang, Binh-Son Hua, and Sai-Kit Yeung.
\newblock Riconv++: Effective rotation invariant convolutions for 3d point
  clouds deep learning.
\newblock {\em IJCV}, 130(5):1228--1243, 2022.

\bibitem{zhao2019rotation}
Chen Zhao, Jiaqi Yang, Xin Xiong, Angfan Zhu, Zhiguo Cao, and Xin Li.
\newblock Rotation invariant point cloud classification: Where local geometry
  meets global topology.
\newblock {\em arXiv preprint arXiv:1911.00195}, 2019.

\bibitem{zou2021geometry}
Longkun Zou, Hui Tang, Ke Chen, and Kui Jia.
\newblock Geometry-aware self-training for unsupervised domain adaptation on
  object point clouds.
\newblock In {\em CVPR}, pages 6403--6412, 2021.

\end{thebibliography}


\begin{thebibliography}{10}\itemsep=-1pt

\bibitem{achituve2021self}
Idan Achituve, Haggai Maron, and Gal Chechik.
\newblock Self-supervised learning for domain adaptation on point clouds.
\newblock In {\em WACV}, pages 123--133, 2021.

\bibitem{cardace2023self}
Adriano Cardace, Riccardo Spezialetti, Pierluigi~Zama Ramirez, Samuele Salti,
  and Luigi Di~Stefano.
\newblock Self-distillation for unsupervised 3d domain adaptation.
\newblock In {\em WACV}, pages 4166--4177, 2023.

\bibitem{chen2022devil}
Ronghan Chen and Yang Cong.
\newblock The devil is in the pose: Ambiguity-free 3d rotation-invariant
  learning via pose-aware convolution.
\newblock In {\em CVPR}, pages 7472--7481, 2022.

\bibitem{chen2024local}
Yiyang Chen, Lunhao Duan, Shanshan Zhao, Changxing Ding, and Dacheng Tao.
\newblock Local-consistent transformation learning for rotation-invariant point
  cloud analysis.
\newblock In {\em CVPR}, pages 5418--5427, 2024.

\bibitem{Deng_2021_ICCV}
Congyue Deng, Or Litany, Yueqi Duan, Adrien Poulenard, Andrea Tagliasacchi, and
  Leonidas~J. Guibas.
\newblock Vector neurons: A general framework for so(3)-equivariant networks.
\newblock In {\em ICCV}, pages 12200--12209, 2021.

\bibitem{huang2021metasets}
Chao Huang, Zhangjie Cao, Yunbo Wang, Jianmin Wang, and Mingsheng Long.
\newblock Metasets: Meta-learning on point sets for generalizable
  representations.
\newblock In {\em CVPR}, pages 8863--8872, 2021.

\bibitem{li2021closer}
Feiran Li, Kent Fujiwara, Fumio Okura, and Yasuyuki Matsushita.
\newblock A closer look at rotation-invariant deep point cloud analysis.
\newblock In {\em ICCV}, pages 16218--16227, 2021.

\bibitem{liang2022point}
Hanxue Liang, Hehe Fan, Zhiwen Fan, Yi Wang, Tianlong Chen, Yu Cheng, and
  Zhangyang Wang.
\newblock Point cloud domain adaptation via masked local 3d structure
  prediction.
\newblock In {\em ECCV}, pages 156--172. Springer, 2022.

\bibitem{luo2022equivariant}
Shitong Luo, Jiahan Li, Jiaqi Guan, Yufeng Su, Chaoran Cheng, Jian Peng, and
  Jianzhu Ma.
\newblock Equivariant point cloud analysis via learning orientations for
  message passing.
\newblock In {\em CVPR}, pages 18932--18941, 2022.

\bibitem{qin2019pointdan}
Can Qin, Haoxuan You, Lichen Wang, C-C~Jay Kuo, and Yun Fu.
\newblock Pointdan: A multi-scale 3d domain adaption network for point cloud
  representation.
\newblock {\em NeurIPS}, 32, 2019.

\bibitem{su2022svnet}
Zhuo Su, Max Welling, Matti Pietik{\"a}inen, and Li Liu.
\newblock Svnet: Where so (3) equivariance meets binarization on point cloud
  representation.
\newblock In {\em 3DV}, pages 547--556. IEEE, 2022.

\bibitem{wang2024progressive}
Zicheng Wang, Zhen Zhao, Yiming Wu, Luping Zhou, and Dong Xu.
\newblock Progressive classifier and feature extractor adaptation for
  unsupervised domain adaptation on point clouds.
\newblock In {\em ECCV}, pages 37--55. Springer, 2024.

\bibitem{wei2022learning}
Xin Wei, Xiang Gu, and Jian Sun.
\newblock Learning generalizable part-based feature representation for 3d point
  clouds.
\newblock {\em NeurIPS}, 35, 2022.

\bibitem{you2021prin}
Yang You, Yujing Lou, Ruoxi Shi, Qi Liu, Yu-Wing Tai, Lizhuang Ma, Weiming
  Wang, and Cewu Lu.
\newblock Prin/sprin: On extracting point-wise rotation invariant features.
\newblock {\em IEEE TPAMI}, 44(12):9489--9502, 2021.

\bibitem{zhang2022riconv}
Zhiyuan Zhang, Binh-Son Hua, and Sai-Kit Yeung.
\newblock Riconv++: Effective rotation invariant convolutions for 3d point
  clouds deep learning.
\newblock {\em IJCV}, 130(5):1228--1243, 2022.

\bibitem{zou2021geometry}
Longkun Zou, Hui Tang, Ke Chen, and Kui Jia.
\newblock Geometry-aware self-training for unsupervised domain adaptation on
  object point clouds.
\newblock In {\em CVPR}, pages 6403--6412, 2021.

\end{thebibliography}
}

\end{document}

\title{Rotation-Adaptive Point Cloud Domain Generalization via Intricate Orientation Learning \\
—— Supplementary Material —— }

\author{{Bangzhen~Liu,~Chenxi~Zheng,~Xuemiao~Xu,~Cheng Xu,~Huaidong~Zhang, \\ and~Shengfeng~He,~\IEEEmembership{Senior Member,~IEEE}}

\thanks{ Bangzhen Liu,~Chenxi~Zheng, and~Xuemiao~Xu are with the School of Computer Science and Engineering, South China University of Technology, Guangzhou, China. E-mail: liubz.scut@gmail.com,~cszcx@mail.scut.edu.cn, and~xuemx@scut.edu.cn.}
\thanks{ Cheng Xu is with the Centre for Smart Health, The Hong Kong Polytechnic University, Hong Kong. E-mail: cschengxu@gmail.com}
\thanks{ Huaidong Zhang is with the School of Future Technology, South China University of Technology, Guangzhou, China. E-mail: huaidongz@scut.edu.cn.}
\thanks{ Shengfeng He is with the School of Computing and Information Systems, Singapore Management University, Singapore. E-mail: shengfenghe@smu.edu.sg.}
}

\markboth{IEEE Transactions on Pattern Analysis and Machine Intelligence}%
{Shell \MakeLowercase{\textit{Liu et al.}}: Rotation-Adaptive Point Cloud Domain Generalization via Intricate Orientation Learning}

\maketitle

\IEEEdisplaynontitleabstractindextext

\IEEEpeerreviewmaketitle

\section{More Experimental Results} \label{sec1}

It is worth noting that the three sub-datasets used in PointDA are all category-wise imbalanced, as shown in Table~\ref{table:dataset}, which indicates that the micro-average precision score (\textit{Acc.}) reported by previous studies is inappropriate to assess the generalizability of cross-domain classification. In the main paper, we instead report the results of PointDA in the form of the macro-average precision score (\textit{Avg.}) for a more convincing evaluation. We also report the extra evaluations in the form of \textit{Acc.} in Table~\ref{tab:pointda10_acc} for reference. Our method still outperforms all the competitors in the average metric over the six cross-domain tasks.

\begin{table}[h]
    \caption{{Number of samples for each category in PointDA~\cite{qin2019pointdan}.}}
    \vspace{-2ex}
    \label{table:dataset}
    \scriptsize
    % \begin{center}
    \setlength{\tabcolsep}{0.05cm}{
      \resizebox{0.48\textwidth}{!}{
        \begin{tabular}{c|c|c|c|c|c|c|c|c|c|c|c}
          \hline  & Tub & Bed & Shelf & Case & Chair & Lamp & Monit. & Plant & Sofa & Table & Total\\
          \hline 
          ModelNet & 106 & 515 & 572 & 200 & 889 & 124 & 465 & 240 & 680 & 392 & 4183 \\
          ShapeNet & 599 & 167 & 310 & 1076 & 4612 & 1620 & 762 & 158 & 2198 & 5876 & 17378 \\
          ScanNet & 98 & 329 & 464 & 650 & 2578 & 161 & 210 & 88 & 495 & 1037 & 6110 \\
          \hline
          \end{tabular}
      }
     }
    % \end{center}
    \vspace{-2mm}
  \end{table}

\noindent\textbf{Evaluation on Aligned Dataset.} {We additionally implement our method under the traditional aligned data scenario, where the rotation only happens on the z-axis. In this case, we adapt our intricate orientation mining approach to specifically identify the most intricate orientations along the z-axis. 
The comparisons with state-of-the-art 3DDG methods are shown in Table~\ref{tab:align}, where the results of competitors are directly borrowed from their papers. Our method surpasses the baselines on all six tasks, demonstrating its effectiveness. The proposed orientation-aware contrastive training enables the model to gain a more comprehensive understanding of point clouds from various challenging perspectives, thereby enhancing the generalizability of the learned features. We notice that our method is slightly inferior on M$\to$S* and S$\to$S*. Since the orientational shift is our major concern, we do not have a special design for capturing geometric information under self-occlusions. However, in this case, our method still outperforms the two 3DDG methods on three out of the six tasks, while achieving the best average accuracy. Furthermore, the experimental results also reveal the presence of rotational shifts in the aligned data scenes, demonstrating the potential of our method for solving this problem.}

\begin{table}[h] % table for OSDA setting on Office31
    \caption{Comparison of the \textit{Acc.} ($\%$) under the 3D domain generalization setting. The best records are marked in \textbf{bold}.}
    \label{tab:align} 
    \vspace{-3ex}
    \small
    \begin{center}
    \setlength{\tabcolsep}{0.1cm}{
    \resizebox{0.48\textwidth}{!}{
    \begin{tabular}{c|c|c|c|c|c|c|c}
    \hline

    Methods
    &{M$\to$S} & {M$\to$S*} & {S$\to$M} & {S$\to$S*} & {S*$\to$M} & {S*$\to$S} & {Avg}\\

    \hline
    Supervised                      &{93.9} &{78.4} &{96.2} &{78.4} &{96.2} &{93.9} &{89.5}\\
    w/o Adapt                       &{83.3} &{43.8} &{75.5} &{42.5} &{63.8} &{64.2} &{62.2}\\
    \hline
    {Metasets~\cite{huang2021metasets}} &\tb{86.0} &{52.3} &{67.3} &{42.1} &{69.8} &{69.5} &{64.5}\\
    {PDG~\cite{wei2022learning}}        &{85.6} &\tb{57.9} &{73.1} &{50.0} &{70.3} &{66.3} &{67.2}\\
    % \hline
    {Ours}                              &{83.8} &{46.0} &\tb{83.2} &{45.5} &\tb{76.4} &\tb{70.3} &\tb{67.5}\\
    \hline

    \end{tabular}

    }
    }
    \end{center}
    \vspace{-3mm}
\end{table}

\begin{table*}[h] % table for OSDA setting on Office31
    \caption{Comparison of the micro-average precision score \textit{Acc.}~($\%$) under the orientation-aware 3D domain generalization setting. The top 2 records are marked in \bc{red} and \rc{blue}.}
    \label{tab:pointda10_acc} 
    \vspace{-3ex}
    \small
    \begin{center}
    \setlength{\tabcolsep}{0.35cm}{
    \resizebox{1\textwidth}{!}{
    \begin{tabular}{c|c|c|c|c|c|c|c|c}
    \hline

    {Methods}
    &Type &{M$\to$S} & {M$\to$S*} & {S$\to$M} & {S$\to$S*} & {S*$\to$M} & {S*$\to$S} & {AVG} \\

    % \cline{2-19}
    % \cline{14-19}  
    % \cmidrule $\pm$ r{2-13}
    % \cmidrule $\pm$ r{14-19}
    \hline
    Supervised                            &\multirow{2}{*}{-}            &{86.6 $\pm$ 6.3}    &{69.6 $\pm$ 3.2}    &{88.4 $\pm$ 14.7}   &{69.6 $\pm$ 3.2}    &{88.4 $\pm$ 14.7}  &{86.6 $\pm$ 6.3}    &{81.5}\\
    w/o Adapt                             &          &{57.9 $\pm$ 15.1}   &{28.7 $\pm$ 5.1}    &{54.1 $\pm$ 8.6}   &{28.8 $\pm$ 4.7}    &{43.0 $\pm$ 4.8}  &{42.3 $\pm$ 6.0}    &{42.5}\\
    \hline   
    VN~\cite{Deng_2021_ICCV}              &\multirow{2}{*}{RE}             &{70.5 $\pm$ 0.0} &{30.6 $\pm$ 0.0} &{66.7 $\pm$ 0.0} &{32.0 $\pm$ 0.0} &{39.4 $\pm$ 0.0} &{44.8 $\pm$ 0.0}   &{47.3} \\  
    SVN~\cite{su2022svnet}               &            &{66.8 $\pm$ 0.6} &{32.3 $\pm$ 0.4}   &{62.0 $\pm$ 0.5} &{30.0 $\pm$ 0.6}    &{38.0 $\pm$ 0.9} &{42.2 $\pm$ 1.1}   &{45.7} \\ 
    EOMP~\cite{luo2022equivariant}        &              &{61.4 $\pm$ 0.8} &{28.1 $\pm$ 0.3}    &{60.5 $\pm$ 0.6} &{37.0 $\pm$ 0.7}  &{27.9 $\pm$ 0.8} &{37.2 $\pm$ 0.9}    &{42.0} \\
    
    \hline
    SPRIN~\cite{you2021prin}              &\multirow{5}{*}{RI}             &{68.2 $\pm$ 0.4} &{30.1 $\pm$ 0.6}   &{71.8 $\pm$ 0.6} &{30.4 $\pm$ 0.6}    &{46.8 $\pm$ 0.6} &{49.3 $\pm$ 0.5}  &{49.4}\\
    RIPCA~\cite{li2021closer}              &            &{70.3 $\pm$ 1.2} &{33.0 $\pm$ 0.7}   &{70.4 $\pm$ 0.9} &{39.1 $\pm$ 1.3}    &\rc{49.9 $\pm$ 1.6} &{50.6 $\pm$ 2.2}  &\rc{52.2}\\
    RIConv++~\cite{zhang2022riconv}        &              &{28.8 $\pm$ 0.6} &{14.2 $\pm$ 0.5}   &{55.1 $\pm$ 0.7} &{38.9 $\pm$ 0.5}    &{34.8 $\pm$ 0.7} &{47.3 $\pm$ 0.5}  &{36.5}\\
    PaRI~\cite{chen2022devil}              &             &{36.1 $\pm$ 0.0} &{29.3 $\pm$ 0.3}   &{51.8 $\pm$ 0.8} &\bc{44.8 $\pm$ 0.4}    &{43.3 $\pm$ 0.9} &{49.4 $\pm$ 0.1} &{42.5} \\
    LocoTrans~\cite{chen2024local}              &                  &\bc{76.7 $\pm$ 0.0}   &{34.5 $\pm$ 0.3}    &\rc{74.3 $\pm$ 0.4}  &\rc{43.6 $\pm$ 0.2}  &{41.6 $\pm$ 0.6} &{41.5 $\pm$ 0.0}    &{52.0}\\
    \hline
    PointDAN~\cite{qin2019pointdan}       &\multirow{5}{*}{DA}              &{59.8 $\pm$ 15.1}   &{29.5 $\pm$ 4.0}    &{55.2 $\pm$ 6.9}   &{24.0 $\pm$ 4.8}    &{38.0 $\pm$ 4.8}  &{47.4 $\pm$ 6.1}    &{42.3}\\
    DefRec~\cite{achituve2021self}        &              &{57.2 $\pm$ 13.3}   &{33.1 $\pm$ 4.6}    &{54.4 $\pm$ 8.0}    &{33.1 $\pm$ 4.4}   &{38.8 $\pm$ 6.5}  &{48.2 $\pm$ 5.7}    &{44.1}\\
    GAST~\cite{zou2021geometry}           &              &{27.7 $\pm$ 4.2}   &{7.0 $\pm$ 0.6}     &{40.8 $\pm$ 2.6}   &{5.8 $\pm$ 0.8}     &{30.7 $\pm$ 1.5}  &{50.7 $\pm$ 3.6}     &{27.1}\\
    MLSP~\cite{liang2022point}            &              &{66.5 $\pm$ 15.5}   &{32.8 $\pm$ 4.3}    &{59.7 $\pm$ 5.1}   &{30.0 $\pm$ 6.4}    &{46.3 $\pm$ 5.0}  &{52.2 $\pm$ 5.8}     &{47.9}\\
    SDDA~\cite{cardace2023self}           &              &{65.0 $\pm$ 14.5}  &\rc{37.8 $\pm$ 3.4}     &{61.4 $\pm$ 5.4}    &{40.1 $\pm$ 4.1}   &{40.7 $\pm$ 6.3}  &\rc{53.3 $\pm$ 6.4}     &{49.7}\\
    PCFEA~\cite{wang2024progressive}     &                   &{62.0 $\pm$ 13.6}   &{9.3 $\pm$ 0.2}   &{42.7 $\pm$ 8.7}   &{43.1 $\pm$ 4.0}   &{47.1 $\pm$ 4.0}   &\bc{54.0 $\pm$ 4.6}    &{43.0}    \\
    \hline
    {Metasets~\cite{huang2021metasets}}   &\multirow{3}{*}{DG}               &{53.9 $\pm$ 1.4} &\bc{40.3 $\pm$ 0.9}   &{32.2 $\pm$ 12.3}  &{33.5 $\pm$ 1.7} &{24.5 $\pm$ 4.6}    &{39.8 $\pm$ 10.0} &{37.4}\\
    {PDG~\cite{wei2022learning}}          &             &{25.4 $\pm$ 29.5}   &{21.2 $\pm$ 18.0}   &{38.4 $\pm$ 18.5}   &{8.1 $\pm$ 3.2}   &{30.3 $\pm$ 4.8}   &{29.7 $\pm$ 12.0}    &{25.5}\\
    % \hline
    {Ours}                                &              &\rb{70.8 $\pm$ 2.0}   &{37.2 $\pm$ 1.2}    &\bb{80.7 $\pm$ 0.6}   &{34.0 $\pm$ 1.0}     &\bb{50.0 $\pm$ 2.5}   &{47.1 $\pm$ 3.2}     &\bb{53.3}  \\
    \hline

    \end{tabular}

    }
    }
    \end{center}
    \vspace{-3mm}
\end{table*}

\begin{figure}[h]
    % \flushleft
    \centering
    \subfloat[Ablation of $\lambda_{oc}$]{%[b]{0.45\textwidth}
        \label{fig:plot_lambda_oc}
        \includegraphics[width=0.22\textwidth]{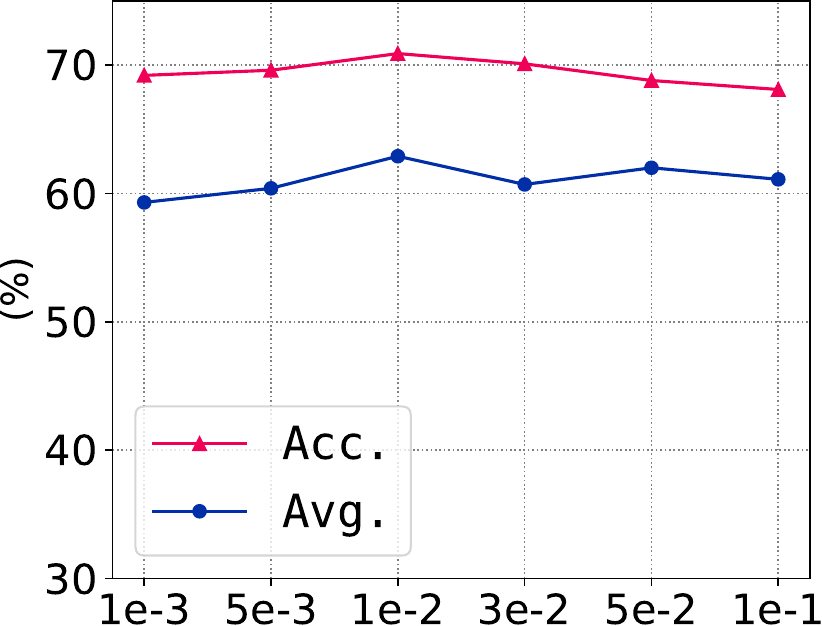}
    }
    % \hspace{2mm}
    \subfloat[Ablation of $\lambda_{ms}$]{%[b]{0.45\textwidth}
        \label{fig:plot_lambda_ms}
        \includegraphics[width=0.22\textwidth]{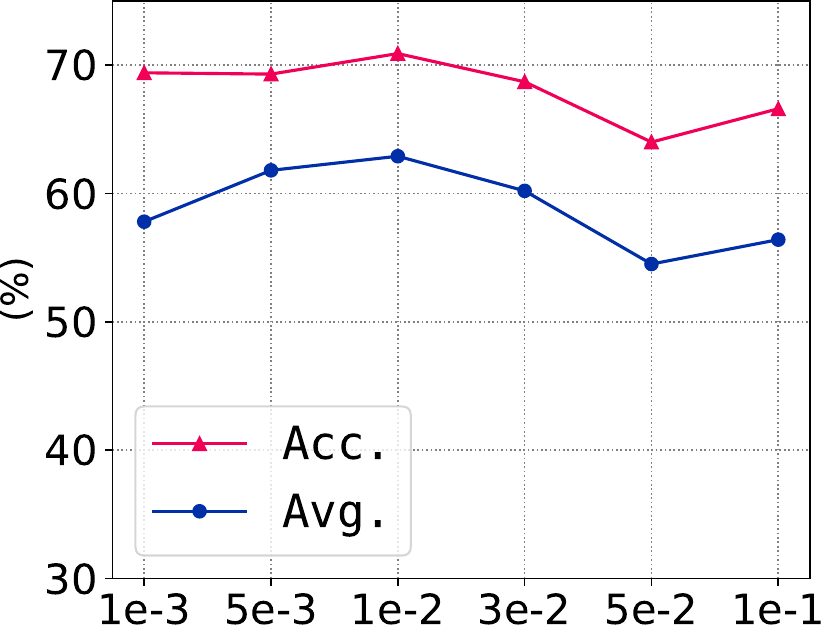}
    }
    % \vspace{-2mm}
    \caption{The curves of performance \wrt varying $\lambda_{oc}$ and $\lambda_{ms}$.}
    \label{fig:ablation}
    % \vspace{-5.5mm}
\end{figure}

\noindent\textbf{Analysis of Hyper-parameter Sensitivity.} 
We evaluate the effects of varying $\lambda_{oc}$ and $\lambda_{ms}$, by changing the value while keeping the other frozen as 0.1. As Fig.~\ref{fig:ablation}(a) and Fig.~\ref{fig:ablation}(b) show, $\lambda_{oc}$ is insensitive across a large range, while larger $\lambda_{ms}$ may slightly decrease the performance of our model. According to the variation of performance curves, we choose $\lambda_{oc}=0.01$ and $\lambda_{ms}=0.01$ as the model setting in our main paper.

\begin{figure}[h]
    % \flushleft
    \centering
    \subfloat[M$\to$S]{%[b]{0.45\textwidth}
        \label{fig:m2s_loss}
        \includegraphics[width=0.23\textwidth]{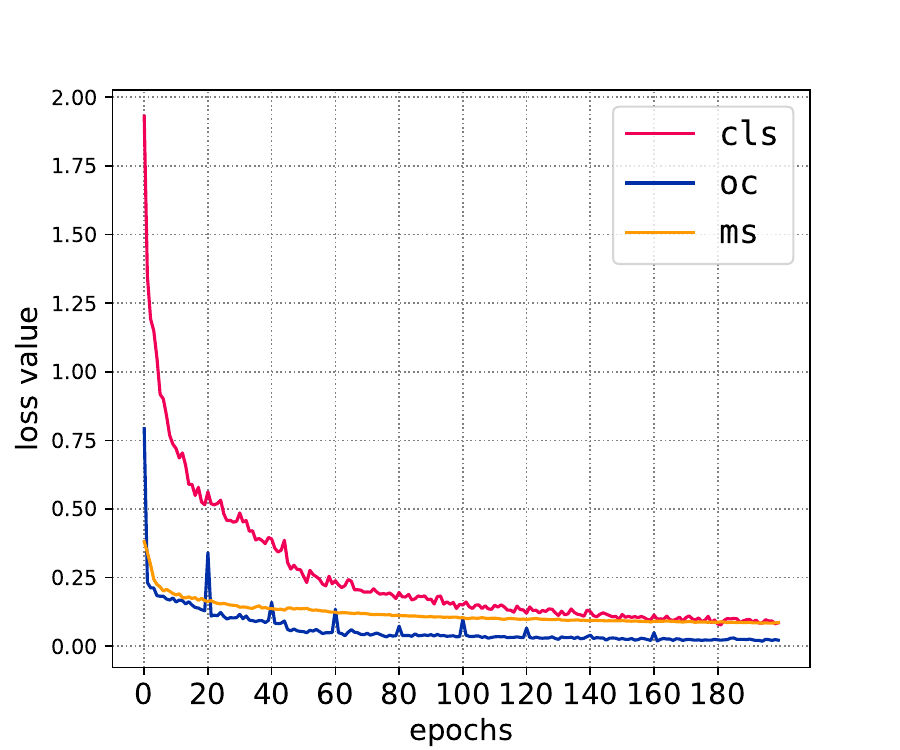}
    }
    % \hspace{2mm}
    \subfloat[M$\to$S*]{%[b]{0.45\textwidth}
        \label{fig:m2ss_loss}
        \includegraphics[width=0.23\textwidth]{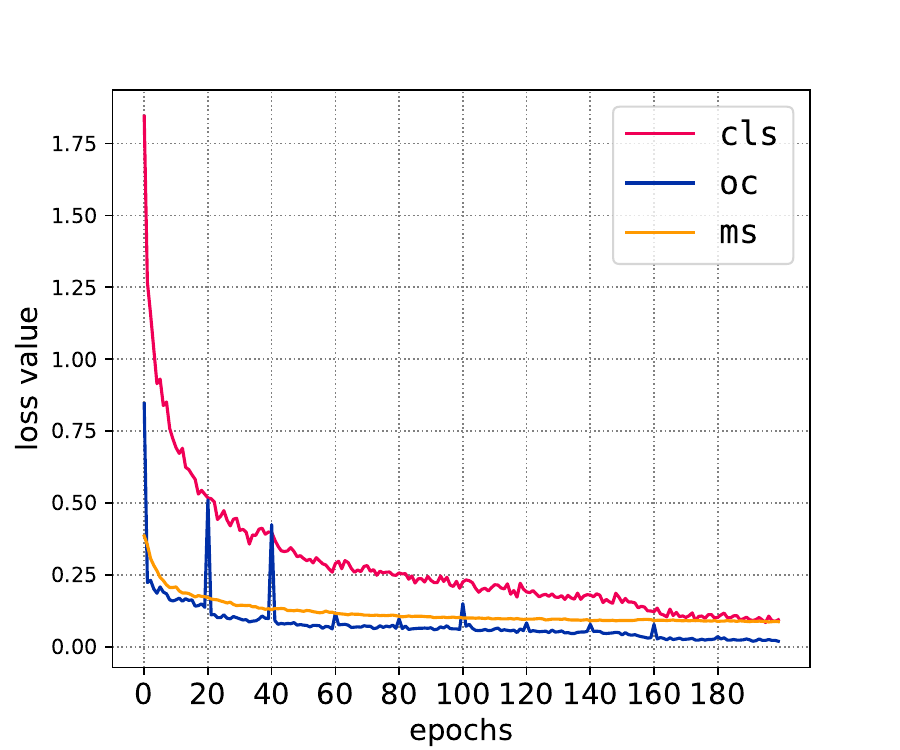}
    }
    \vspace{-2ex}
    \caption{The training curves (\ie, $L_{cls}$, $L_{oc}$, and $L_{ms}$) on M$\to$S (a) and M$\to$S* (b).}
    \label{fig:plot}
    \vspace{-3mm}
\end{figure}

\noindent\textbf{Analysis of Training Stability.} 
We plot the curves of the proposed orientation consistency loss and the marginal separation loss over the training stage to demonstrate the convergence of our intricate orientational learning. As Fig.~\ref{fig:plot}(a) and Fig.~\ref{fig:plot}(b) show, all the losses gradually decrease and converge to a convincing degree. The blue curves are the orientation consistency loss, which periodically bursts every 20 epochs. This is due to the update of the intricate orientation set, which gradually adapts the model to all the intricate orientations. At the end of the training stage, the amplification tends to be stable, indicating the consistency of the object towards various rotations.

\noindent\textbf{Analysis of Time Complexity.} {We report the computational costs of training/testing one batch of data in milliseconds for different compared methods in Table~\ref{tab:complexity}. The results are obtained by accumulating the running times within a single training/testing epoch and calculating the mean value w.r.t. one batch.} Due to the process of diversifying the intricate orientation set, our method introduces extra computational costs in the training phase. Nonetheless, our method yields the best performance among these methods while achieving the second-best inferencing speed, which is more efficient than the other RE and RI methods that require extra time-consuming modules for practical applications.

\begin{table}[h]
    \caption{Time statistics (ms) of training/testing on one batch of data.}
    \label{tab:complexity}
    \vspace{-3ex}
    \small
    \begin{center}
    \setlength{\tabcolsep}{0.3cm}{
    \resizebox{0.4\textwidth}{!}{
      \begin{tabular}{c|c|c|c|c} 
        \hline 
        Methods & Type & Avg. & $T_{train}$& $T_{test}$ \\
        \hline
        VN~\cite{Deng_2021_ICCV} & RE & 41.3 & 808 & 27.9 \\
        SPRIN~\cite{you2021prin} & RI & 43.9 & 1551 & 370.2 \\
        RIPCA~\cite{li2021closer} & RI & 46.6 & 717 & 23.3 \\
        MLSP~\cite{liang2022point} & DA & 43.2 & 825 & 36.5 \\
        SDDA~\cite{cardace2023self} & DA & 43.1 & \bf{567} & \bf{12.0} \\
        \hline
        Ours & DG & \bf{49.6} & 2114 & 14.5 \\
        \hline
        \end{tabular}
    }
    }
    \end{center}
    \vspace{-3mm}
  \end{table}

\section{Extra Visualizations and Analysis} \label{sec2}

\begin{figure*}[h]
    % \flushleft
    \centering
    \subfloat{\label{fig:0}
        \begin{minipage}[b]{1.0\textwidth}\centering
            \includegraphics[width=0.95\textwidth]{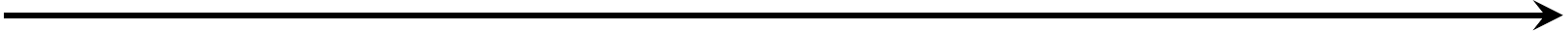} 
            \\
            \includegraphics[width=0.9\textwidth]{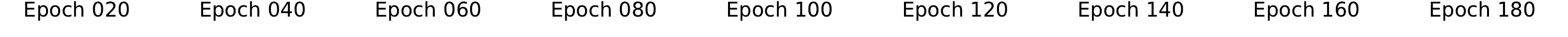} 
            \\
            \includegraphics[width=0.9\textwidth]{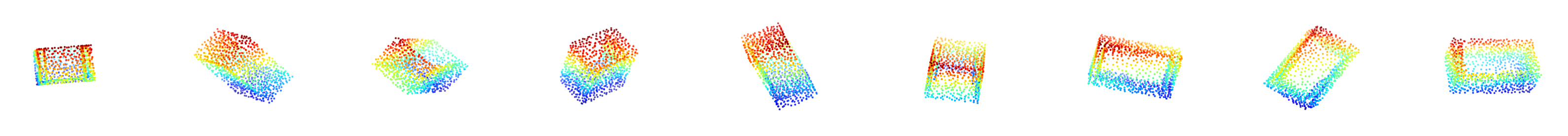} 
            \\
            \vspace{-3mm}
            \includegraphics[width=1.0\textwidth]{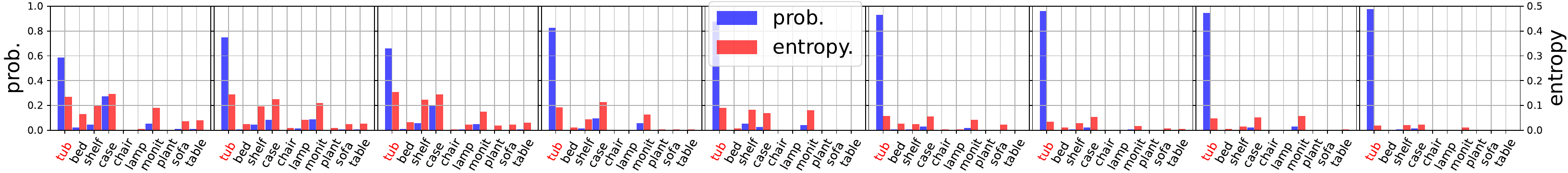}
        \end{minipage}
    }\vspace{-3mm}

    \subfloat{\label{fig:1}
        \begin{minipage}[b]{1.0\textwidth}\centering
            \includegraphics[width=0.9\textwidth]{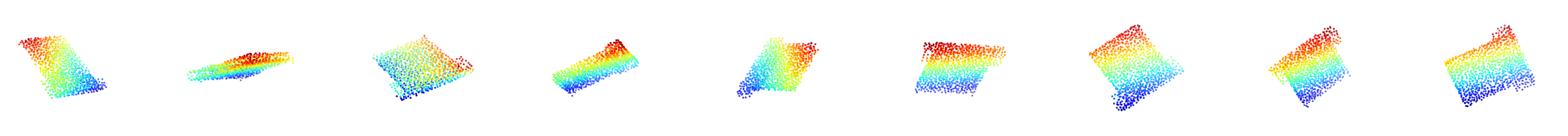} 
            \\
            \vspace{-3mm}
            \includegraphics[width=1.0\textwidth]{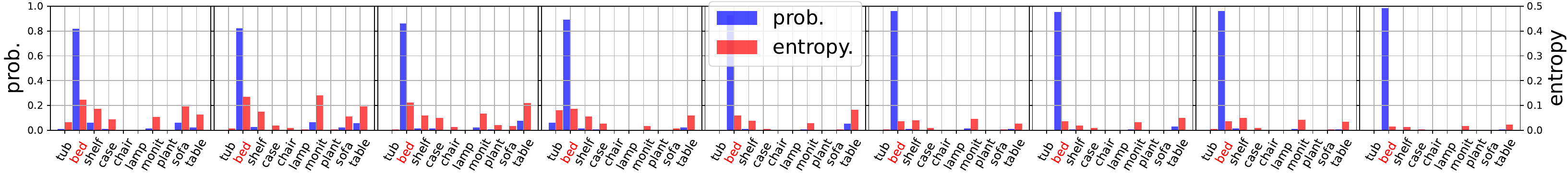}
        \end{minipage}
    }\vspace{-3mm}

    \subfloat{\label{fig:2}
        \begin{minipage}[b]{1.0\textwidth}\centering
            \includegraphics[width=0.9\textwidth]{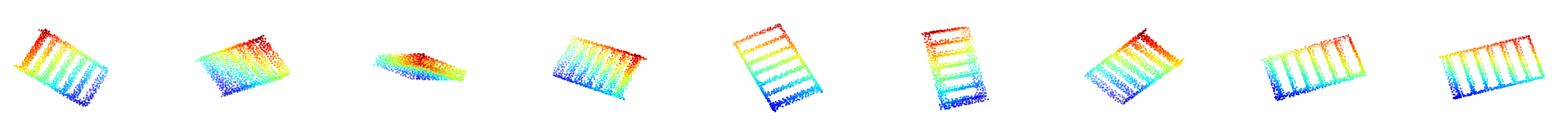} 
            \\
            \vspace{-3mm}
            \includegraphics[width=1.0\textwidth]{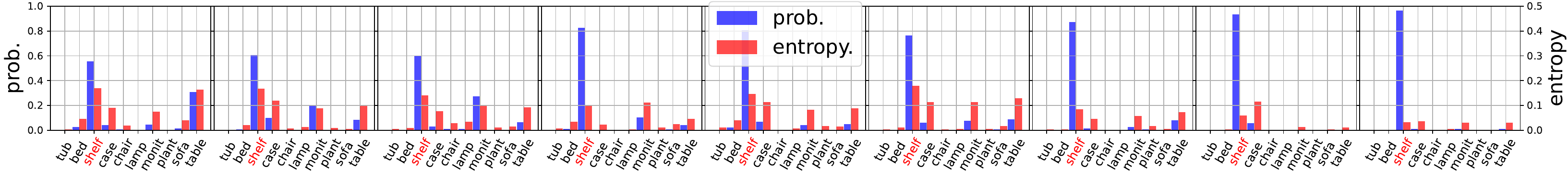}
        \end{minipage}
    }\vspace{-3mm}

    \subfloat{\label{fig:4}
        \begin{minipage}[b]{1.0\textwidth}\centering
            \includegraphics[width=0.9\textwidth]{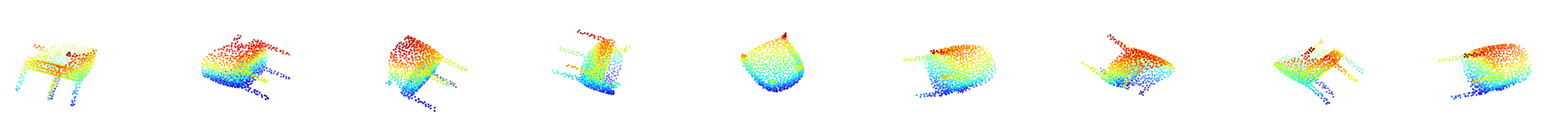} 
            \\
            \vspace{-3mm}
            \includegraphics[width=1.0\textwidth]{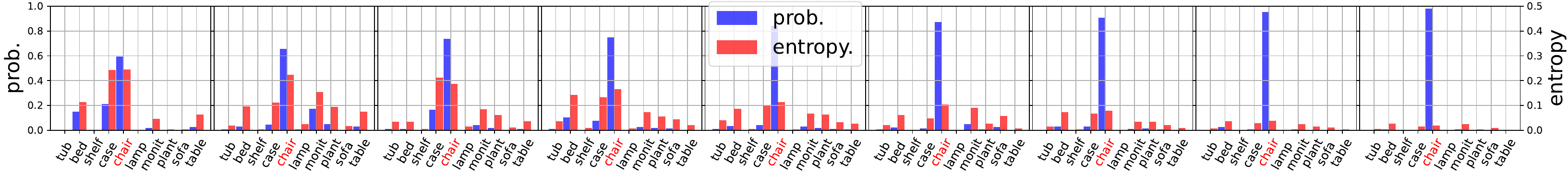}
        \end{minipage}
    }\vspace{-3mm}

    \subfloat{\label{fig:6}
        \begin{minipage}[b]{1.0\textwidth}\centering
            \includegraphics[width=0.9\textwidth]{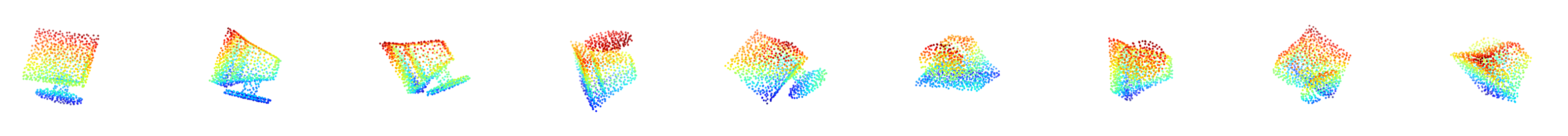} 
            \\
            \vspace{-3mm}
            \includegraphics[width=1.0\textwidth]{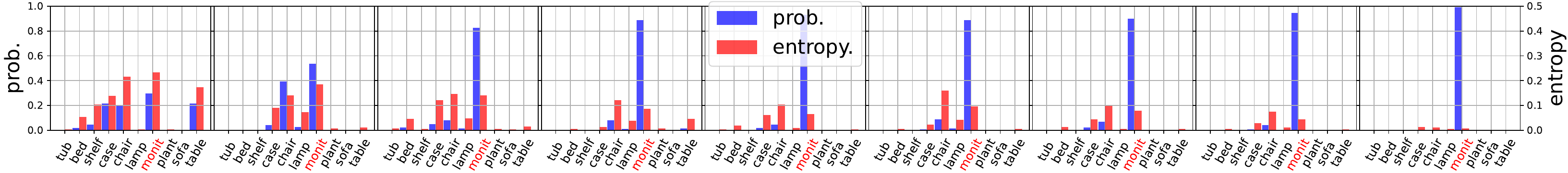}
        \end{minipage}
    }\vspace{-3mm}

    \subfloat{\label{fig:7}
        \begin{minipage}[b]{1.0\textwidth}\centering
            \includegraphics[width=0.9\textwidth]{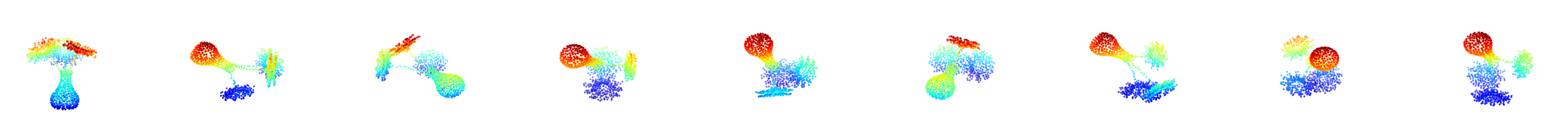} 
            \\
            \vspace{-3mm}
            \includegraphics[width=1.0\textwidth]{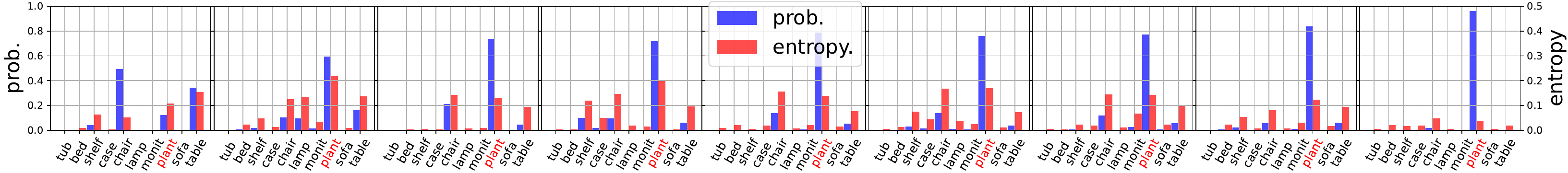}
        \end{minipage}
    }\vspace{-3mm}

    \subfloat{\label{fig:8}
        \begin{minipage}[b]{1.0\textwidth}\centering
            \includegraphics[width=0.9\textwidth]{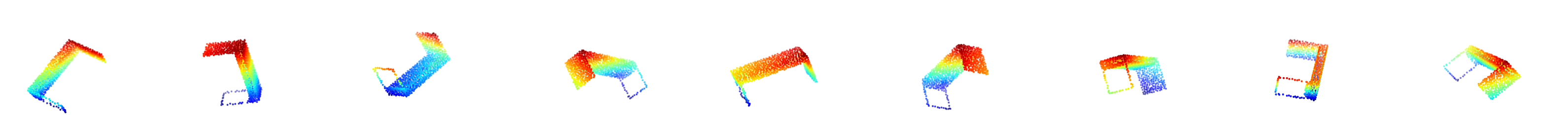} 
            \\
            \vspace{-3mm}
            \includegraphics[width=1.0\textwidth]{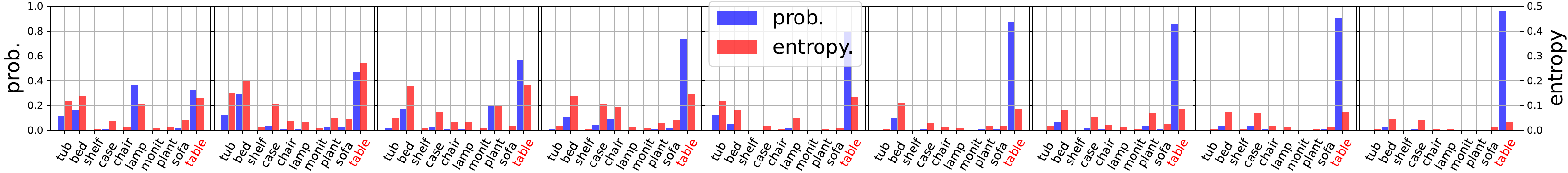}
        \end{minipage}
    }

    \caption{Visualization of the learned intricate orientation series on ModelNet (M). Each row of point cloud sequence records the transformation of the point cloud's poses after augmention by its corresponding intricate rotational angle during the training procedure. The interval of recording is 20 epoch. The statistic measurements at every records, including the predicted probability and rotational consistency, are presented beneath each point cloud sequence. The category which the current point cloud belongs to is marked in \bc{red}.}
    \label{fig:intricat_angle}
    % \vspace{-6mm}
\end{figure*}

\noindent\textbf{The Learned Intricate Augmented Samples.} {In Fig.~\ref{fig:intricat_angle}, we select several point clouds and provide visualizations of how their intricate orientations evolve during training. We trained the model on ModelNet and optimized the intricate set on the testing set every 20 epochs. Each row of the point cloud sequence shows the current pose of the given point cloud augmented by its corresponding intricate orientation at that specific epoch. 
Beneath each sequence, we also visualize the distribution of predicted probabilities and the consistency of prediction over different testing orientations. 
Specifically, for each point cloud, we obtain the predicted probabilities of its 64 testing variants $P = {\{P_a|P_a = \left[p^1_a, ..., p^C_a\right]\}}^A_{a=1}$, where $A=64$ is the number of testing orientation series and $C=10$ is the number of categories. 
The visualized probabilities' distribution $P_m$ is calculated by averaging the predictions over the 64 testing rotation series, such that $P_m = \left[\frac{1}{A}\sum_{j=1}^{A}p^1_j, ..., \frac{1}{A}\sum_{j=1}^{A}p^C_j\right]$. 
To evaluate the predicted consistency, we adopt the entropy as the metric and calculate the consistency $Ent_m$ over the 64 predicted probabilities by 
\begin{equation*}
  Ent_m = \left[\frac{1}{A}\sum_{j=1}^{A}p^1_j log p^1_j, ..., \frac{1}{A}\sum_{j=1}^{A}p^C_j log p^C_j\right].
\end{equation*}
As the number of training epochs increases, both the confidence and output consistency of the model are enhanced. For samples located near the decision boundaries, such as row 6 and row 9, learning with intricate orientation mining could significantly alleviate the ambiguity of learned features, thereby producing a more robust and generalizable classifier for downstream tasks.
}

\begin{figure*}[h]
    % \flushleft
    \centering
    \subfloat[Metasets]{%[b]{0.45\textwidth}
        \label{fig:cm1_m2s}
        % \centering
        \includegraphics[width=0.24\textwidth]{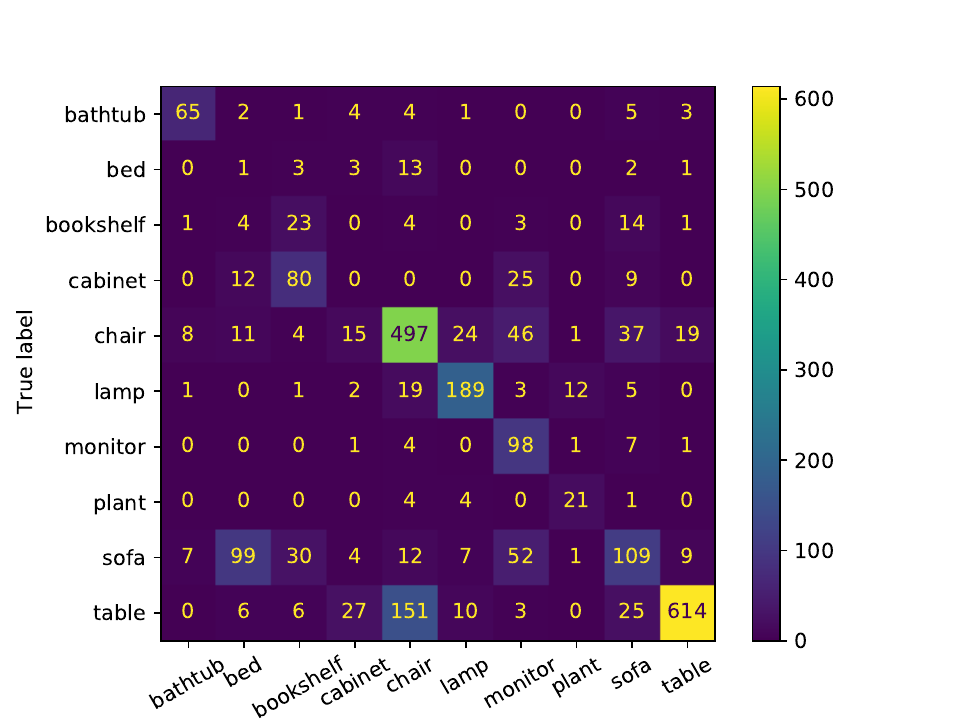}
        % \vspace{10mm}
    }
    \subfloat[PDG]{%[b]{0.45\textwidth}
        \label{fig:cm2_m2s}
        % \centering
        \includegraphics[width=0.24\textwidth]{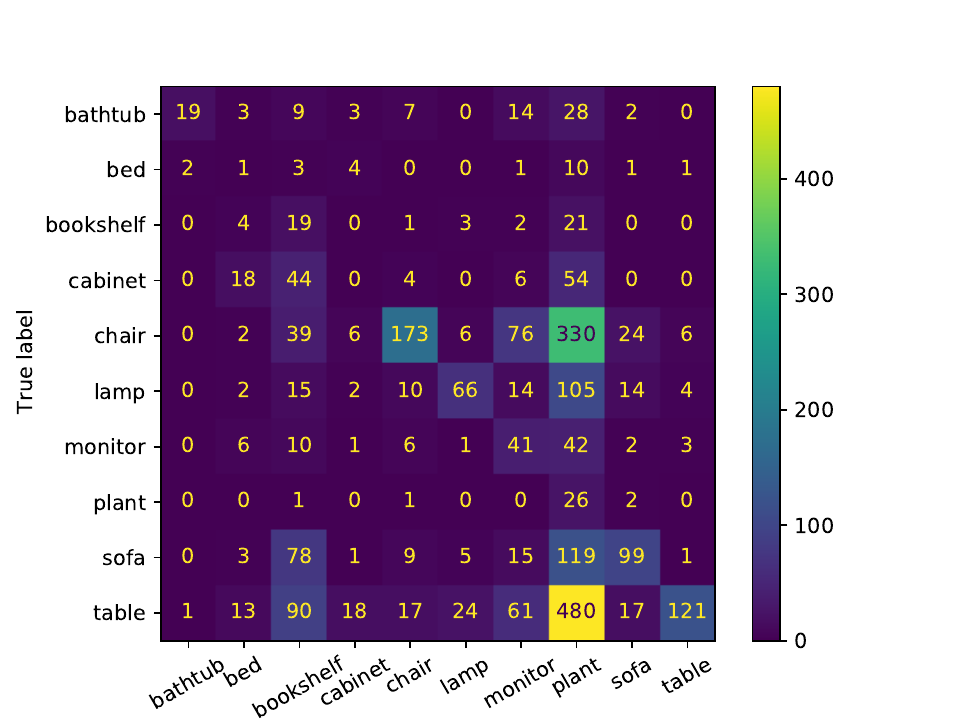}
        % \vspace{10mm}
    }
    \subfloat[Ours]{%[b]{0.45\textwidth}
        \label{fig:cm3_m2s}
        % \centering
        \includegraphics[width=0.24\textwidth]{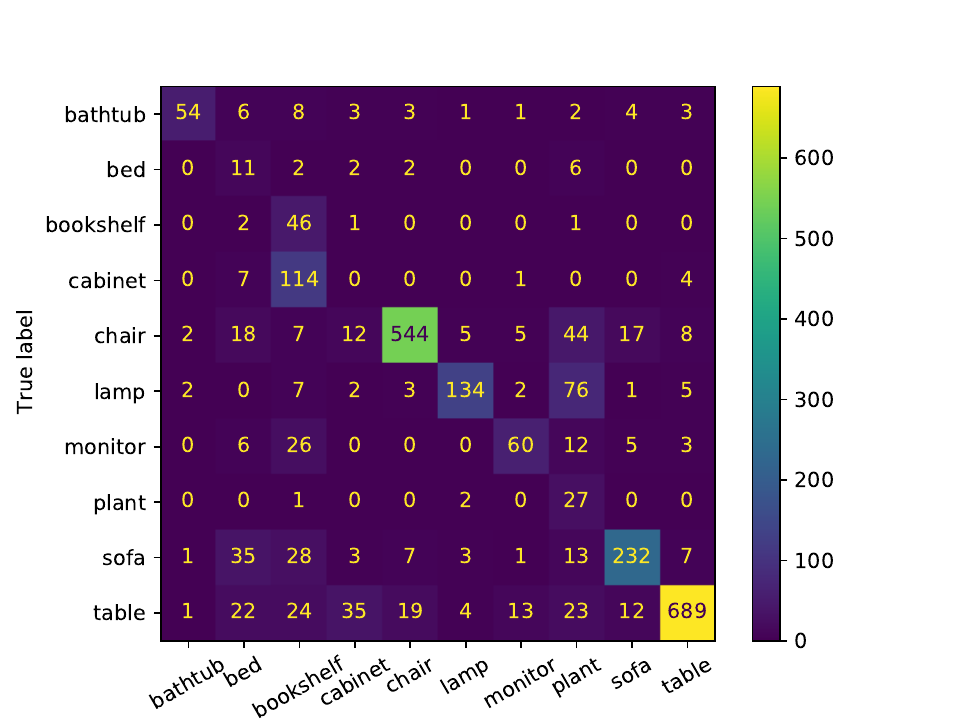}
        % \vspace{10mm}
    }
    \vspace{-2ex}
    \caption{The confusion matrices of Metaset, PDG, and our method on M$\to$S. Zoom in for details.}
    \vspace{-3ex}
    \label{fig:visualization_m2s}
\end{figure*} 
% \vspace{-4mm}
\begin{figure*}[h]
    % \flushleft
    \centering
    \subfloat[Metasets]{%[b]{0.45\textwidth}
        \label{fig:cm1_s2m}
        % \centering
        \includegraphics[width=0.24\textwidth]{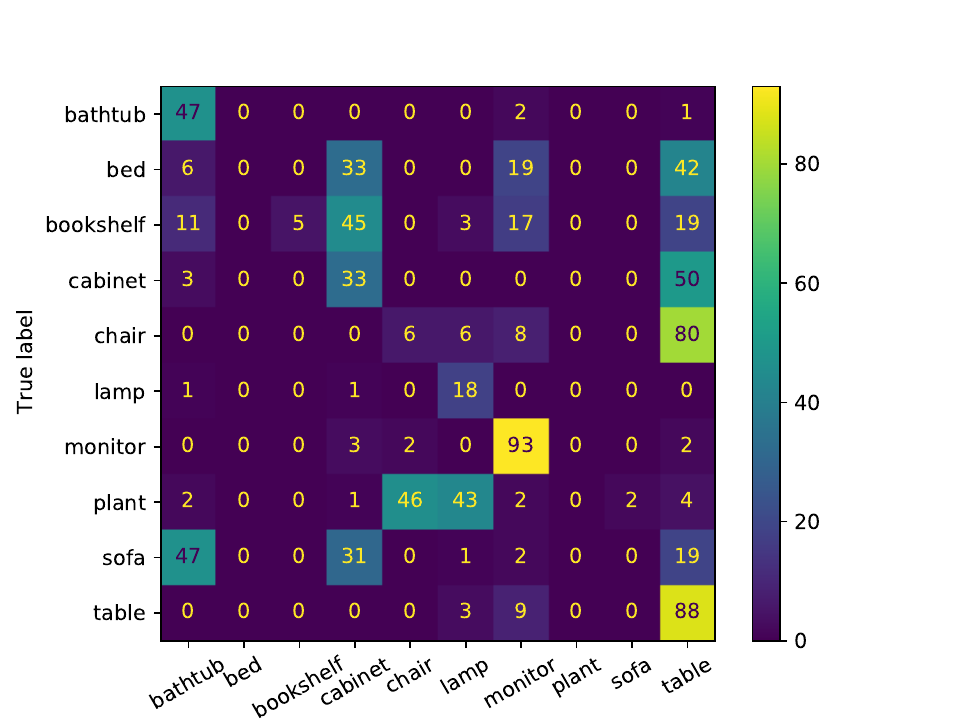}
        % \vspace{10mm}
    }
    \subfloat[PDG]{%[b]{0.45\textwidth}
        \label{fig:cm2_s2m}
        % \centering
        \includegraphics[width=0.24\textwidth]{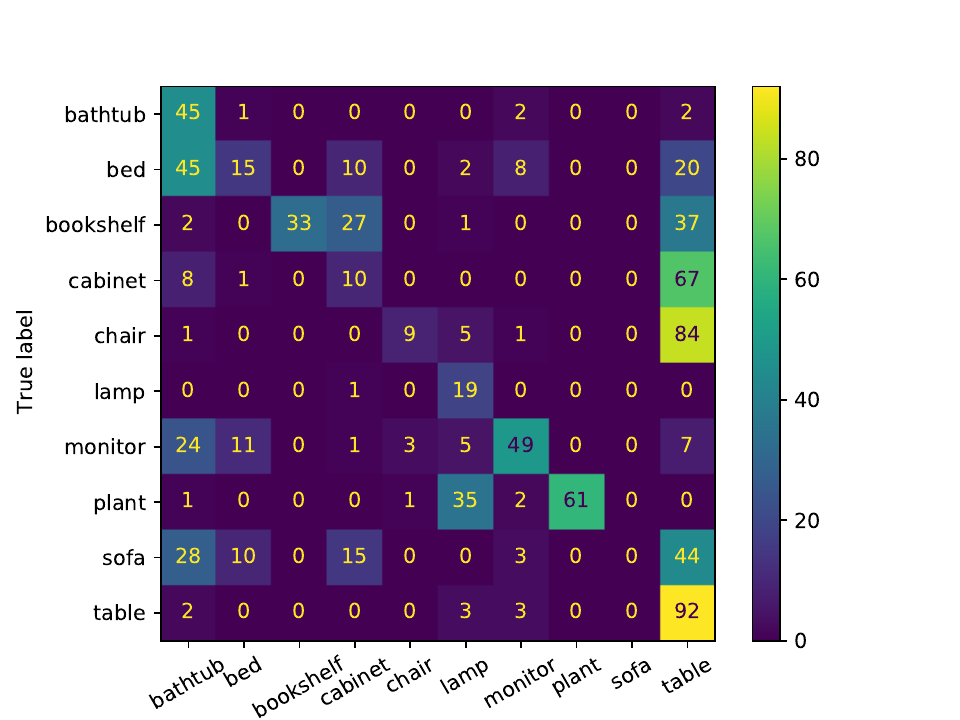}
        % \vspace{10mm}
    }
    \subfloat[Ours]{%[b]{0.45\textwidth}
        \label{fig:cm3_s2m}
        % \centering
        \includegraphics[width=0.24\textwidth]{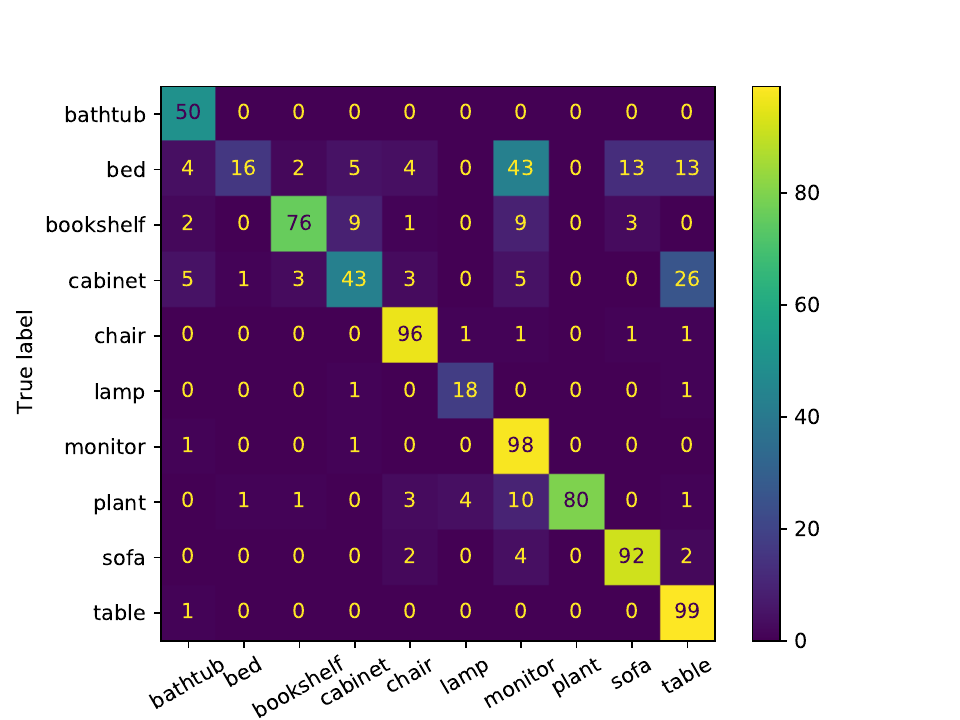}
        % \vspace{10mm}
    }
    \vspace{-2ex}
    \caption{The confusion matrices of Metaset, PDG, and our method on S$\to$M. Zoom in for details.}
    \label{fig:visualization_s2m}
\end{figure*}

\noindent\textbf{Confusion Matrices.} {we provide the evaluation results for Metasets~\cite{huang2021metasets}, PDG~\cite{wei2022learning}, and our method in the form of confusion matrix on the target domain. ShapeNet is a dataset whose samples are highly imbalanced across different categories, while ModelNet is much more balanced. The confusion matrices of the three approaches are shown in  Fig.~\ref{fig:visualization_m2s} and Fig.~\ref{fig:visualization_s2m}. Compared with the other two 3D domain generalization methods, our method has more compact confusion matrices under the orientation shift. For M$\to$S, both our method and Metaset are separated relatively well while PDG has much inaccurate classification on class "Plant". The inner reason is that the part-based feature utilized by PDG may encounter confusing local expressions, such as the plane of the table and the bottom of a potted plant. For S$\to$M, our method achieves more balanced and concise results. We observe that the sample of class "monitor" is much easier to misclassify into "bed" due to the similar plane structure of their surface. Similar trends happen for the categories "table" and "cabinet", which have less discriminative features in the view of shape. }

{In summary, the single shape cannot serve as a discriminative representation in some cases. This is the limitation of shape representation under the orientation shift since there are a lot of objects whose shapes are similar but belong to different categories. In this case, extra visual (\eg, texture or color), linguistic information, or spatial cues are important to provide complement representation, which may benefit the problem of cross-domain generalization under orientation shift. We will plan to investigate the function of these features in our future work.  }

\section{Gradient of the rotation parameters} \label{sec3}
In this section, we provide detailed calculations about the optimizable parameters $\Theta$ concerning a given model $F$. Considering the objective of optimizing $\Theta$ within a standard classification task, we have the following objective:
\begin{equation}
  \hat{\Theta} = \mathop{\arg\max_{\Theta}}L(w_{opt}, \hat{P}, y), 
\end{equation}
where $w_{opt}$ is the freeze parameter of $F$, $(\hat{P}, y)$ are the augmented point cloud and label:
\begin{equation}
  \begin{split}
    ~\hat{P}&=f(\hat{\Theta}, P)  \\
    &=R_{\theta_{x}}\cdot R_{\theta_{y}}\cdot R_{\theta_{z}}\cdot P.
  \end{split}
\end{equation}
According to the chain rules, the gradient of $\hat{\Theta}$ is calculated by:
\begin{equation}
  \begin{split}
  \frac{\partial L}{\partial \hat{\Theta}} &= \frac{\partial L}{\partial \hat{P}} \frac{\partial \hat{P}}{\partial \hat{\Theta}} \\
  &=\frac{\partial L}{\partial \hat{P}}
  \left(
  \frac{\partial R_{\theta_{x}}}{\partial \theta_{x}}
  R_{\theta_{y}}
  R_{\theta_{z}} \quad
  R_{\theta_{x}}
  \frac{\partial R_{\theta_{y}}}{\partial \theta_{y}}
  R_{\theta_{z}} \quad
  R_{\theta_{x}}
  R_{\theta_{y}}
  \frac{\partial R_{\theta_{z}}}{\partial \theta_{z}}
  \right)P,
  \end{split}
\end{equation}
where 
\begin{equation}
  \begin{split}
R_{\theta_{x}} = 
\begin{pmatrix}
  1 & 0 & 0 \\
  0 & \cos\theta_{x} & -\sin\theta_{x} \\
  0 & \sin\theta_{x} & \cos\theta_{x} 
\end{pmatrix},
\\
R_{\theta_{y}} = 
\begin{pmatrix}
  \cos\theta_{y} & 0 & \sin\theta_{y} \\
  0 & 1 & 0 \\
  -\sin\theta_{y} & 0 & \cos\theta_{y}
\end{pmatrix},
\\
R_{\theta_{z}} = 
\begin{pmatrix}
  \cos\theta_{z} & -\sin\theta_{z} & 0  \\
  \sin\theta_{z} & \cos\theta_{z} & 0 \\
  0 & 0 & 0
\end{pmatrix},
\end{split}
\end{equation}
and 
\begin{equation}
    \begin{split}
\frac{\partial R_{\theta_{x}}}{\partial \theta_{x}} = 
\begin{pmatrix}
  0 & 0 & 0 \\
  0 & -\sin\theta_{x} & -\cos\theta_{x} \\
  0 & \cos\theta_{x} & -\sin\theta_{x} 
\end{pmatrix}, \\
\frac{\partial R_{\theta_{y}}}{\partial \theta_{y}} = 
\begin{pmatrix}
  -\sin\theta_{y} & 0 & \cos\theta_{y} \\
  0 & 0 & 0 \\
  -\cos\theta_{y} & 0 & -\sin\theta_{y}
\end{pmatrix}, \\
\frac{\partial R_{\theta_{z}}}{\partial \theta_{z}} = 
\begin{pmatrix}
  -\sin\theta_{z} & -\cos\theta_{z} & 0  \\
  \cos\theta_{z} & -\sin\theta_{z} & 0 \\
  0 & 0 & 1
\end{pmatrix}.
\end{split}
\end{equation}

\section{Theoretical Analysis for Rotation-Adaptive Point Cloud Domain Generalization} \label{sec4}

In this section, we provide theoretical proof demonstrating how orientational consistency functions to bridge the domain gap, analyzed from the perspective of mutual information reduction.

Let $X\!=\!(U, V)$ represent a 3D point cloud, where $U$ corresponds to orientation-dependent variables and $V$ to orientation-independent variables. In our case, we assume that the ranges of $U$ and $V$ remain consistent across domains.
For $X_s\!\sim\!p_\mathrm{src}(x)$, where $p_\mathrm{src}(x)$ denotes the source domain data distribution, the marginal distributions \wrt $U_s$ and $V_s$ are expressed by:
\begin{equation}
 p_\mathrm{src}(u)=\int p_\mathrm{src}(x) \mathrm{d}v, \quad p_\mathrm{src}(v)=\int p_\mathrm{src}(x) \mathrm{d}u.
\end{equation}
Considering the data distribution $X_a\!\sim\!p_\mathrm{aug}(x)$ after augmentation, where each sample is assumed to be uniformly sampled \wrt orientations, the marginal distributions \wrt $U_a$ and $V_a$ are given by:
\begin{equation}
 p_\mathrm{aug}(u)=\mathcal{U}(\mathcal{D}_{U_a}), \quad p_\mathrm{aug}(v)=p_\mathrm{src}(v),
\end{equation}
where $\mathcal{U}(\cdot)$ denotes a uniform distribution over the measurable domain $\mathcal{D}_{U_a}$ of ${U_a}$. For simplicity, the subscript of $U_a$ in $\mathcal{D}_{U_a}$ is omitted without causing ambiguity in the subsequent analysis. In this work, we adopt the proposed orientation-aware contrastive learning framework to approximately achieve this, where ${U_a}$ is represented by Euler angles and $\mathcal{D}_{U}:=[-\pi, \pi)^3$.

Based on the definition of joint entropy, the entropy of $p_\mathrm{src}(x)$ can be expressed in terms of its marginal entropies \wrt $U_s$ and $V_s$, along with an additional term presenting the mutual information between these two components:
\begin{equation}
\begin{aligned}
    \H(X_s) 
 =& \H(U_s)+\H(V_s)-\I(U_s;V_s) \\
 =& \mathbb{E}_{u\sim p_\mathrm{src}(u)}[-\log p_\mathrm{src}(u)] + \mathbb{E}_{v\sim p_\mathrm{src}(v)}[-\log p_\mathrm{src}(v)] \\
    &- \mathbb{E}_{x\sim p_\mathrm{src}(x)}\log \frac{p_\mathrm{src}(x)}{p_\mathrm{src}(u)p_\mathrm{src}(v)},
\end{aligned}
\end{equation}
where $\I(U_s;V_s)$ represents the mutual information between $U_s$ and $V_s$ in $p_\mathrm{src}(x)$. 
Since $p_\mathrm{aug}(u)$ follows a uniform distribution and $U_a$ and $V_a$ of $p_\mathrm{aug}(x)$ are independent under this setting, the entropy of $p_\mathrm{aug}(x)$ is given by $\I_\mathrm{aug}(U_a;V_a)\!=\!0$, and the entropy of $p_\mathrm{aug}(u)$ corresponds to the measure of $\mathcal{D}_{U}$, denoted as $m(\mathcal{D}_{U})$. 
Thus, the entropy of $p_\mathrm{aug}(x)$ can be simplified as follows:
\begin{equation}
\begin{aligned}
    \H(X_a) &= \H(U_a)+\H(V_a) \\
    &= \log m(\mathcal{D}_{U}) + \mathbb{E}_{v\sim p_\mathrm{aug}(v)}[-\log p_\mathrm{aug}(v)],
\end{aligned}
\end{equation}
where $m(\mathcal{D}_{U})\!=\!(2\pi)^3$ in our case.

We use the KL divergence to quantify the distributional shift between the source and the target distribution.
For any $X_t\!\sim\!p_\mathrm{tgt}(x)$, where $p_\mathrm{tgt}(x)$ represents the target domain distribution, the KL divergence between $p_\mathrm{tgt}(x)$ and $p_\mathrm{src}(x)$ (or $p_\mathrm{aug}(x)$) can be computed once the cross-entropy between them is known. 
However, directly calculating the cross-entropy between $p_\mathrm{tgt}(x)$ and $p_\mathrm{src}(x)$ (or $p_\mathrm{aug}(x)$) is intractable, and it is often treated as an optimization objective to minimize. Notably, the cross-entropy between $p_\mathrm{tgt}(x)$ and $p_\mathrm{src}(x)$ (or $p_\mathrm{aug}(x)$) shares the same upper bound, as the samples $X_s$, $X_a$, and $X_t$ all share the same dimensionality:
\begin{equation}
    \sup_{p_\mathrm{src}}{\H(p_\mathrm{tgt}, p_\mathrm{src})} = \sup_{p_\mathrm{aug}}{\H(p_\mathrm{tgt}, p_\mathrm{aug})} = \log ({m(\mathcal{D}_U) \times m(\mathcal{D}_V)}).
\end{equation}
Here, $\mathcal{D}_V$ is the measurable domain of $V_s$, $V_a$, and $V_t$.
It is straightforward to prove that $\H_\mathrm{aug}(X_a) > \H_\mathrm{src}(X_s)$, as the mutual information is non-negative and entropy reaches its upper bound when the distribution is uniform.
Therefore, the relation between the upper bound of the KL divergence from $p_\mathrm{tgt}(x)$ to $p_\mathrm{src}(x)$ and from $p_\mathrm{tgt}(x)$ to $p_\mathrm{aug}(x)$ can be expressed as:
\begin{equation}
\begin{aligned}
    \sup_{p_\mathrm{src}}{\KL(p_\mathrm{tgt}||p_\mathrm{src})} &= \sup_{p_\mathrm{src}}{\H(p_\mathrm{tgt};p_\mathrm{src})} - \sup_{p_\mathrm{src}}{\H(X_\mathrm{s})} \\
    &> \sup_{p_\mathrm{aug}}{\H(p_\mathrm{tgt};p_\mathrm{aug})} - \sup_{p_\mathrm{aug}}{\H(X_\mathrm{a})} \\
    &= \sup_{p_\mathrm{aug}}{\KL(p_\mathrm{tgt}||p_\mathrm{aug})}. \label{eq:ieq}
\end{aligned}
\end{equation}
As revealed in Eq.~\ref{eq:ieq}, the upper bound of $\KL(p_\mathrm{tgt}||p_\mathrm{aug})$ is consistently lower than $\KL(p_\mathrm{tgt}||p_\mathrm{src})$, demonstrating the effectiveness of orientation invariance in reducing the domain shift under the disturbance of varying rotations. Consequently, the final upper bound of $\KL(p_\mathrm{tgt}||p_\mathrm{aug})$ is formally given as follows:
\begin{equation}
    \sup_{p_\mathrm{aug}}{\KL(p_\mathrm{tgt}||p_\mathrm{aug})} = \log m(\mathcal{D}_V) - \mathbb{E}_{v\sim p_\mathrm{aug}(v)}[-\log p(v)].
\end{equation}

\section{Limitation and Future Work} 
Although our method shows commendable advantages in handling cross-domain orientational shifts, it faces challenges with other complex types of domain shifts, such as heavy occlusions. This is because our framework does not offer an explicit design for tackling these domain shifts. Addressing this limitation, possibly through constructing a more powerful and versatile feature space resilient to multiple domain shifts via self-supervised pre-training, is a goal for future work.

{\small
\bibliographystyle{ieee_fullname}
\bibliography{egbib}
}